\documentclass[10pt,twocolumn,letterpaper]{article}

\usepackage{cvpr}
\usepackage{times}
\usepackage{epsfig}
\usepackage{url}
\usepackage{graphicx}
\usepackage{subfigure} 
\usepackage{multirow}
\usepackage{algorithm}
\usepackage{algorithmic}
\usepackage{amsthm}
\usepackage{amsmath}
\usepackage{enumitem}
\usepackage{amssymb}
\usepackage{bm}
\usepackage{wrapfig}
\usepackage{defs_151115}
\usepackage{algorithm}
\usepackage{algorithmic}
\graphicspath{ {./figs/} }

\usepackage[pagebackref=true,breaklinks=true,letterpaper=true,colorlinks,bookmarks=false]{hyperref}

 \cvprfinalcopy % *** Uncomment this line for the final submission

\def\cvprPaperID{2373} % *** Enter the CVPR Paper ID here

\ifcvprfinal\fi
\begin{document}

%%%%%%%%% TITLE
\title{Generating images with recurrent adversarial networks}

\author{Daniel Jiwoong Im$^1$\\
Montreal Institute for Learning Algorithms\\
University of Montreal\\
{\tt\small imdaniel@iro.umontreal.ca}
% For a paper whose authors are all at the same institution,
% omit the following lines up until the closing ``}''.
% Additional authors and addresses can be added with ``\and'',
% just like the second author.
% To save space, use either the email address or home page, not both
\and
Chris Dongjoo Kim\\
Department of Engineering and Computer Science\\
York University\\
{\tt\small kimdon20@gmail.com}
\and
Hui Jiang\\
Department of Engineering and Computer Science\\
York University\\
{\tt\small hj@cse.yorku.ca}
\and
Roland Memisevic\\
Montreal Institute for Learning Algorithms\\
University of Montreal\\
{\tt\small memisevr@iro.umontreal.ca}
}

\maketitle
%\thispagestyle{empty}

%%%%%%%%% ABSTRACT
\begin{abstract} 
%    Generative adversarial networks are well-known generative models 
%    that are not based on maximum likelihood. Recent studies have shown that these models 
%    are very good at generating ``data like'' samples.
%    In this paper, we propose sequential modeling using generative recurrent 
%    adversarial network. $\cdots$
Gatys et al. (2015) showed that optimizing pixels to match features 
in a convolutional network is a way to render images of high visual quality. 
Unrolling this gradient-based optimization can be thought of as a recurrent 
computation, that creates images by incrementally adding onto a visual ``canvas''.
%similar to the recently introduced DRAW model. 
Inspired by this view we propose a recurrent generative model that 
can be trained using adversarial training. % to generate very good image samples.
In order to quantitatively compare adversarial networks we also propose a 
new performance measure, that is based on letting the generator and discriminator 
of two models compete against each other. 
\end{abstract}

%%%%%%%%% BODY TEXT
\section{Introduction}
Generating realistic-looking images has been a long-standing goal in 
machine learning. 
The early motivation for generating images was mainly as a diagnostic tool, 
based on the belief that a good generative model can count as evidence for 
the degree of ``understanding'' that a model has of the visual world 
(see, example, \cite{hinton2006}, \cite{Hyvrinen:2009:NIS:1572513}, 
or \cite{ranzato2013modeling} and references in these).  
More recently, due to immense quality improvements over the last 
two years (for example, \cite{Gregor2015,Denton2015,Radford2015,gatys2015neural}), 
and the successes of discriminative modeling overall, 
image generation has become a goal on its own, with industrial 
applications within close reach. 

Currently, most common image generation models can be roughly categorized into 
two classes: 
The first is based on probabilistic generative models, such 
as the variational autoencoder \cite{Kingma2014vae} and a variety of 
equivalent models introduced at the same time. The idea in these models is 
to train an autoencoder whose latent representation satisfies certain 
distributional properties, which makes it easy to sample from the hidden variables,
as well as from the data distribution (by plugging samples into the decoder). 

The second class of generative models is based on adversarial sampling \cite{Goodfellow2014}.
This approach forgoes the need to encourage a particular latent distribution 
(and, in fact, the use of an encoder altogether), by training a simple feed-forward neural network 
to generate ``data-like'' examples. ``Data-likeness'' is judged by a 
simultaneously trained, but otherwise separate, discriminator neural network. 

For both types of approach, sequential variants were introduced recently, 
which were shown to work much better in terms of visual quality: 
The DRAW network \cite{Gregor2015}, for example, is a sequential version of 
the variational autoencoder, where images are generated by accumulating updates 
into a canvas using a recurrent network. 
%This also allows the use of an attention mechanism which was shown to be crucial for obtaining good samples. 
An example of a sequential adversarial network is the LAPGAN model \cite{Denton2015},
which generates images in a coarse-to-fine fashion, by generating and upsampling in 
multiple steps. 

Motivated by the successes of sequential generation, in this paper, we propose 
a new image generation model based on a recurrent network. 
Similar to \cite{Denton2015}, our model generates an image in a sequence of structurally 
identical steps, but in contrast to that work we do not impose a coarse-to-fine
(or any other) structure on the generation procedure. Instead we let the 
recurrent network learn the optimal procedure by itself. 
In contrast to \cite{Gregor2015}, we obtain very good samples without resorting 
to an attention mechanism and without variational training criteria (such as 
a KL-penalty on the hiddens). 
%The parameters of the network are trained rather than fixed. 

Our model is mainly inspired by a third type of image generation method 
proposed recently by \cite{gatys2015neural}. In this work, the goal is to change the 
texture (or ``style'') of a given reference image by generating a new image that 
matches image features and texture features within the layers of a pretrained 
convolutional network. As shown by \cite{gatys2015neural}, ignoring the style-cost 
in this approach and only matching image features, it is possible 
to render images which are similar to the reference image. 
As we shall show, unrolling the gradient descent based optimization that 
generates the target image yields a recurrent computation, in which 
an ``encoder'' convolutional network extracts images of the current ``canvas''.  
The resulting code and the code for the reference image get fed into a ``decoder'' 
which decides on an update to the ``canvas''. 

This view, along with the successes of trained sequential generation networks, 
suggests that an iterative convolutional network that is trained to 
accumulate updates onto a visual canvas should be good at generating images 
in general, not just those shown as reference images. 
We show in this paper that this indeed is the case. 
%so hybrid 

To evaluate and compare the relative performance of adversarial generative 
models quantitatively, we also introduce a new evaluation scheme based on 
a ``cross-over'' battle between the discriminators and generators of the 
two models.

%The desire to make computers imagine and fantasize data has been a driving force behind research in deep generative models.  Deep generative models evolved rapidly since the introduction of Deep Belief Networks (DBN) \cite{hinton2006}.  
%For example, DBN have spawned Deep Boltzmann Machines \cite{Salakhutdinov2009}.
%Subsequently, a variety of new algorithms were introduced,  such as variational auto-encoders\cite{Kingma2014vae} and generative adversarial networks \cite{Goodfellow2014}.  In this paper, we turn our interests towards a variant of generative adversarial networks. 

%-------------------------------------------------------------------------
\section{Background}
Generative Adversarial Networks (GAN) are built upon the concept of 
a non-cooperative game \cite{Nash1951}, that two networks are trained to play against 
each other. 
The two networks are 
a generative and a discriminative model, $G$ and $D$. 
The generative model generates samples that are hard for the discriminator $D$ 
to distinguish from real data. 
At the same time, the discriminator tries to avoid getting fooled by the 
generative model $G$.

Formally, the discriminative model is a classifier 
$D:\mathbb{R}^M \rightarrow\lbrace0,1\rbrace$ that tries to determine 
whether a given point $\mathbf{x} \in \mathbb{R}^M$ is real or generated data.  
The generative model $G:\mathbb{R}^K \rightarrow \mathbb{R}^M$ generates 
samples $\bx \in \mathbb{R}^M$ that are similar to the data by mapping a 
sample $\bz \in \mathbb{R}^K$ drawn randomly from some prior 
distribution $p(\bz)$ to the data space. 
%For example, when the prior distribution is a Gaussian distribution,
%then a sample is drawn from the Gaussian distribution and it
%gets mapped to $\bx_{sample}$. 
These models can be trained by playing {\em a minmax game} as follows:
\begin{multline}
    \min_{\theta_G} \max_{\theta_D} V(D,G) = \min_{G} \max_{D} \Big[ \mathbb{E}_{\bx \sim p_{\mathcal{D}}}\big[\log D(\bx)\big] \\
    + \mathbb{E}_{\bz \sim p_{\mathcal{G}}}\big[\log \big(1-D(G(\bz))\big)\big] \Big].
    \label{eqn:gan_obj}
\end{multline}
where $\theta_G$ and $\theta_D$ are the parameters of discriminator and generator, respectively.
%where $p_\mathcal{D}$ and $p_\mathcal{G}$ are data and model distributions, respectively.

In practice, the second term in Equation~\ref{eqn:gan_obj} is troublesome 
due to the saturation of $\log \big(1-D(G(\bz))\big)$. This makes
insufficient gradient flow through the generative model $G$ as
the magnitude of gradients get smaller and prevent them from learning.
To remedy the vanishing gradient problem, the objective function in Equation~\ref{eqn:gan_obj} is reformulated 
into two separate objectives:  
\begin{multline}
    \max_{\theta_D} \mathbb{E}_{\bx \sim p_{\mathcal{D}}}\big[\log D(\bx)\big] 
        + \mathbb{E}_{\bz \sim p_{\mathcal{G}}}\big[\log \big(1 - D(G(\bz))\big)\big], \\
    \& \max_{\theta_G}\mathbb{E}_{\bz \sim p_{\mathcal{G}}}\big[\log D\big(G(\bz)\big)\big].
    \label{eqn:gan_pac_obj}
\end{multline}
%\[ V(D,G) = 
%    \begin{cases} 
%        \min_{G} \max_{D} \mathbb{E}_{\bx \sim p_{\mathcal{D}}}\big[\log D(\bx)\big] \\ % \theta_D & \text{if }D(\bx) \text{ predict wrong} \\
%        \max_{G} \mathbb{E}_{\bz \sim p_{\mathcal{G}}}\big[\log D\big(G(\bz)\big)\big] %& \text{if }D(G(\bz)) \text{ predict wrong} \\
%   \end{cases}
%    \label{eqn:gan_pac_obj}
%\]

Although Equation~\ref{eqn:gan_pac_obj} is not the same as Equation~\ref{eqn:gan_obj},
the underlying intuition is the same. Moreover, 
the gradient of generators for the two different objectives are 
always pointing in the same direction and the two objectives have the
same fixed points.

The generating and discriminating procedure are simple. 
We consider a Gaussion prior distribution with zero-mean and unit variance. 
Then, the process of generating an output is simply to pass a sample 
$\bz \sim \mathcal{N}(\bm{\mu}=\bm{0},\bm{\sigma}=\bm{1})$
to the generative model to obtain the sample
$\bx \sim G(\bz; \theta_G)$. Note that the generative model $G$ 
can be a deterministic or a probabilistic model. 
However, only deterministic models have been deployed in the past,
so that $\bx=G(\bz; \theta_G)$. Subsequently, the sample 
can be passed on to the discriminator to predict $D(\bx; \theta_D)$.

After computing the cost in Equation~\ref{eqn:gan_pac_obj},
the model parameters can be updated through backpropagation. 
Due to the two different min-max operators in Equation~\ref{eqn:gan_pac_obj},
the update rule is defined as follows: 
\[ \lbrace \theta_D^\prime, \theta_G^\prime\rbrace \leftarrow
    \begin{cases} 
        \text{Update } \theta_D & \text{if }D(\bx) \text{ predicts wrong} \\
        \text{Update } \theta_D & \text{if }D(G(\bz)) \text{ predicts wrong} \\
        \text{Update } \theta_G & \text{if }D(G(\bz)) \text{ predicts correct}  
   \end{cases}
\]
%as follows:
%\begin{align}
%    \bz &\sim \mathcal{N}(\bm{\mu}=\bm{0},\bm{\sigma}=\bm{1})\\
%    \bx &= G(\bz; \theta)
%\end{align}
%where $G(\cdot)$ represents the selected 
%the generative model, and $\theta$ is the parameter of the generative model G.

Ideally, we would like the generative model to learn a distribution
such that $p_\mathcal{G} = p_{\mathcal{D}}$. This requires the generative
model to be capable of transforming a simple prior distribution
$p(\bz)$ to more complex distributions.
In general, deep neural networks are good candidates as
they are capable of modeling complicated functions and 
they were shown to be effective in previous works \cite{Goodfellow2014,Mirza2014,Gauthier2015,Denton2015}.
%However, note that 
%this min-max continuous game does not neccessarily have a Nash equilibrium solution
%because there are uncountably many possible neural networks.
%to consider due to the parameters of neural network being not in a closed domain.

\begin{figure*}[t]
    \begin{minipage}{0.49\textwidth}
    \includegraphics[width=\columnwidth]{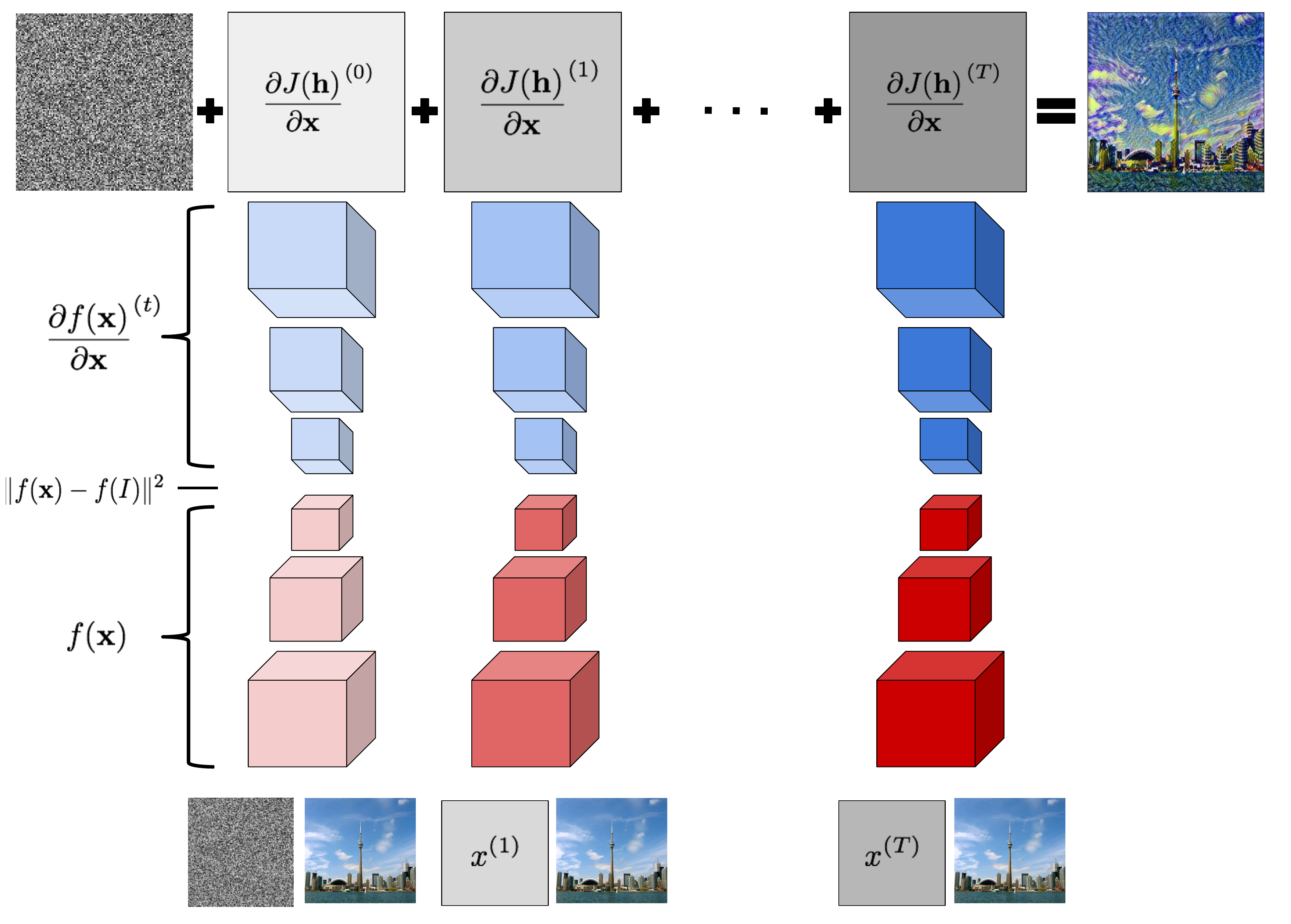}
    \end{minipage}
    \begin{minipage}{0.49\textwidth}
    \includegraphics[width=\columnwidth]{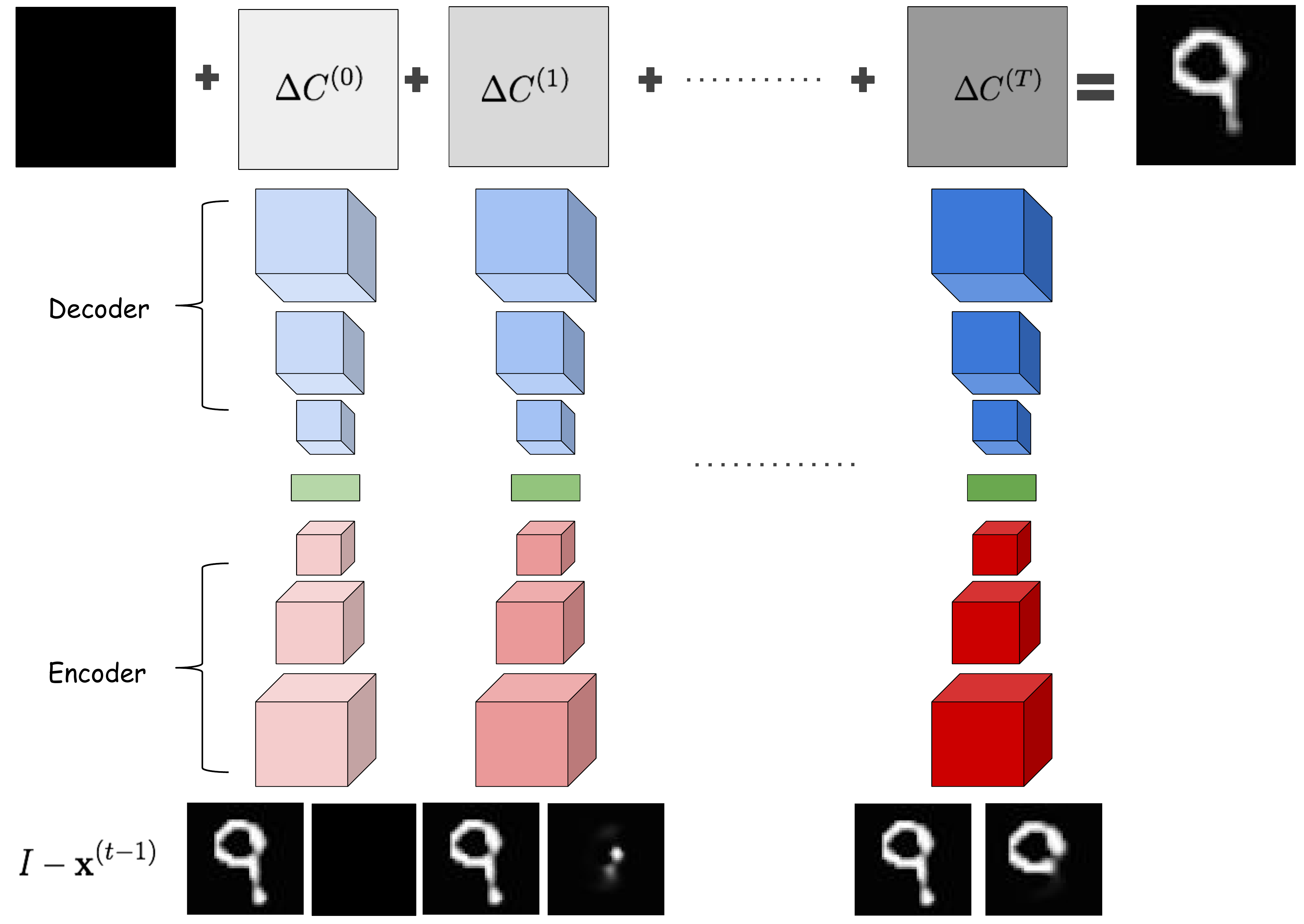}
    \end{minipage}
    \label{fig:drawartify}
    \caption{{\bf Left:} Unrolling the gradient-based optimization of pixels in Gatys et al. {\bf Right:} The DRAW network.}
    \vskip -0.2in
\end{figure*}

Recently, \cite{Radford2015} showed excellent samples of realistic images using 
a fully convolutional neural network as the discriminative model 
and fully deconvolutional 
neural network \cite{Zeiler2011} as the generative model. 
The $l^{th}$ convolutional layer in the discriminative network
takes the form 
%$h^{k^{(l)}}_{j} = f\left(\sum_{j\in M_k} h_j^{l-1} * W^{k^{(l)}} + b^{k^{(l)}}_j \right)$,
\begin{equation}
    h^{k^{(l)}}_{j} = f\left(\sum_{j\in M_k} h_j^{l-1} * W^{k^{(l)}} + b^{k^{(l)}}_j \right),
\end{equation}
and the $l^{th}$ convolutional transpose layer
\footnote{It is more proper to say ``convolutional
transpose operation'' rather than ``deconvolutional" operation. Hence,
we will be using the term ``convolutional transpose'' from now on.} in the generative network takes the form
%$g^{c^{(l)}}_{j} = f\left(\sum_{j\in M_c} g_j^{l-1} \star W^{c^{(l)}} + b^{c^{(l)}}_j \right)$.
\begin{equation}
    g^{c^{(l)}}_{j} = f\left(\sum_{j\in M_c} g_j^{l-1} \star W^{c^{(l)}} + b^{c^{(l)}}_j \right).
\end{equation}
In these equations, $*$ is the convolution operator, 
$\star$ is the convolutional transpose operator, 
$M_j$ is the selection
of inputs from the previous layer (``input maps''), $f$ is an activation function,
and $\lbrace W^{k^{(l)}}, b^{k^{(l)}}_j\rbrace$ and $\lbrace W^{c^{(l)}}, b^{c^{(l)}}_j\rbrace$
are the parameters of the discriminator and generator at layer $l$.
The detailed explanation of convolutional transpose
is explained in the supplementary materials.%~\ref{sec:supp_ct}.

\section{Model}
We propose sequential modeling using GANs on images. Before introducing
our proposed methods, we discuss some of the motivations for our approach.
One interesting aspect of models such as the Deep Recurrent 
Attentive Writer (DRAW) \cite{Gregor2015}
and the Laplacian Generative Adversarial Networks (LAPGAN) \cite{Denton2015} 
is that they 
generate image samples in a sequential process, rather than generating them in one shot. 
Both were shown to outperform their ancestor models, which
are the variational auto-encoder and GAN, respectively. 
The obvious advantage of such sequential models is that
repeatedly generating outputs conditioned on previous states 
simplifies the problem of modeling complicated data distributions 
by mapping them to a sequence of simpler problems.

There is a close relationship between sequential generation 
and {\em Backpropgating to the Input} (BI).
BI is a well-known technique where the goal is to obtain a neural 
network {\em input} that minimizes a given objective function derived 
from the network. 
For example, \cite{gatys2015neural} recently introduced a model 
for stylistic rendering by optimizing the input image to simultaneously 
match higher-layer features of a {\em reference content image} and a non-linear, 
texture-sensitive function of the same features of a {\em reference style image}. 
They also showed that in the absence of the style-cost, this optimization 
yields a rendering of the content image (in a quality that depends on the 
chosen feature layer).

Interestingly, rendering by feature matching in this way is itself 
closely related to DRAW: optimizing a matching cost 
with respect to the input pixels with backprop amounts to first extracting  
the current image features $f_x$ at the chosen layer using a forward path through 
the network (up to that layer). Computing the gradient of the feature reconstruction 
error then amounts to back-propogating the difference $f_x-f_I$ back to the pixels. 
This is equivalent to traversing a ``decoder'' network, defined as the linearized, 
inverse network that computes the backward pass.  
The negative of this derivative is then added into the current version, $\mathbf{x}$, 
of the generated image. We can thus think of image $\mathbf{x}$ as a buffer or 
``canvas'' onto which updates are accumulated 
sequentially (see the left of Figure~\ref{fig:drawartify}). Like in the 
DRAW model, where the updates are computed using a (forward) pass through an encoder 
network, followed by a (backward) pass through a decoder network. 
This approach is almost identical to the DRAW network, except for two 
subtle differences (see, \cite{Gregor2015}): 
(i) in DRAW, the difference between the current image and the image to be 
rendered is used in the forward pass,  whereas here this difference is 
computed in the feature space (after encoding); 
%(ii) in DRAW the decoder is nonlinear;
(ii) DRAW uses a learned, attention-based decoder and encoder rather than (fixed) 
convolutional network. (see the right of Figure~\ref{fig:drawartify}). 
We elaborate on the relationship between the two methods in the supplementary material.

In this work, we explore {\em a generative recurrent adversarial network} 
as an intermediate between DRAW and gradient-based optimization based on a
generative adversarial objective function.%pre-trained network.

\begin{figure*}[t]
    \centering
    \includegraphics[width=0.9\textwidth]{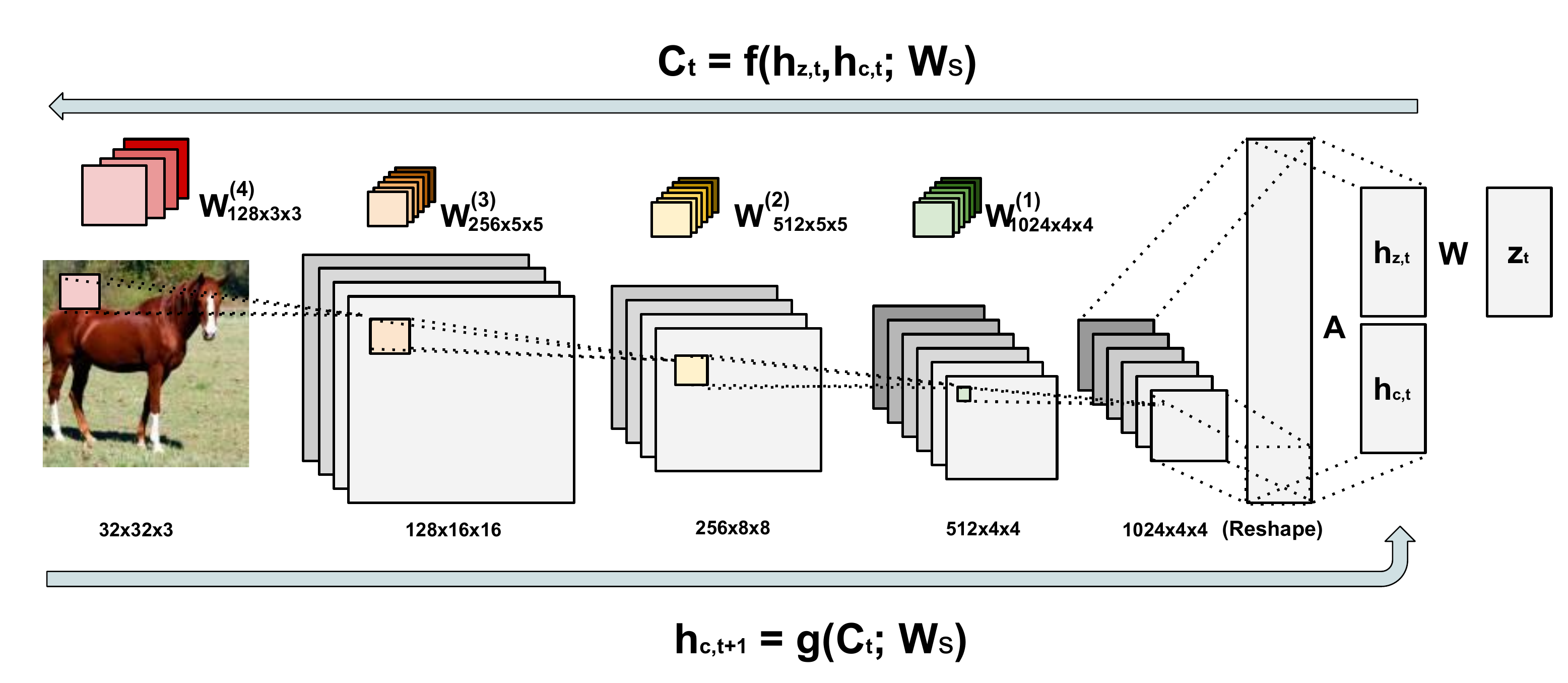}
    \vskip -0.05in
    \caption{Depiction of single time step component of Generative Recurrent Adversarial Networks architecture layed out. 
    (The numbers of the figures are used for modelling CIFAR10 dataset)}
    \label{figs:GRAN}
\vskip -0.2in
\end{figure*} 
\subsection{Generative Recurrent Adversarial Networks}
\begin{figure}[htp]
%\begin{wrapfigure}{r}{0.5\textwidth}
    \centering
    \includegraphics[width=\columnwidth]{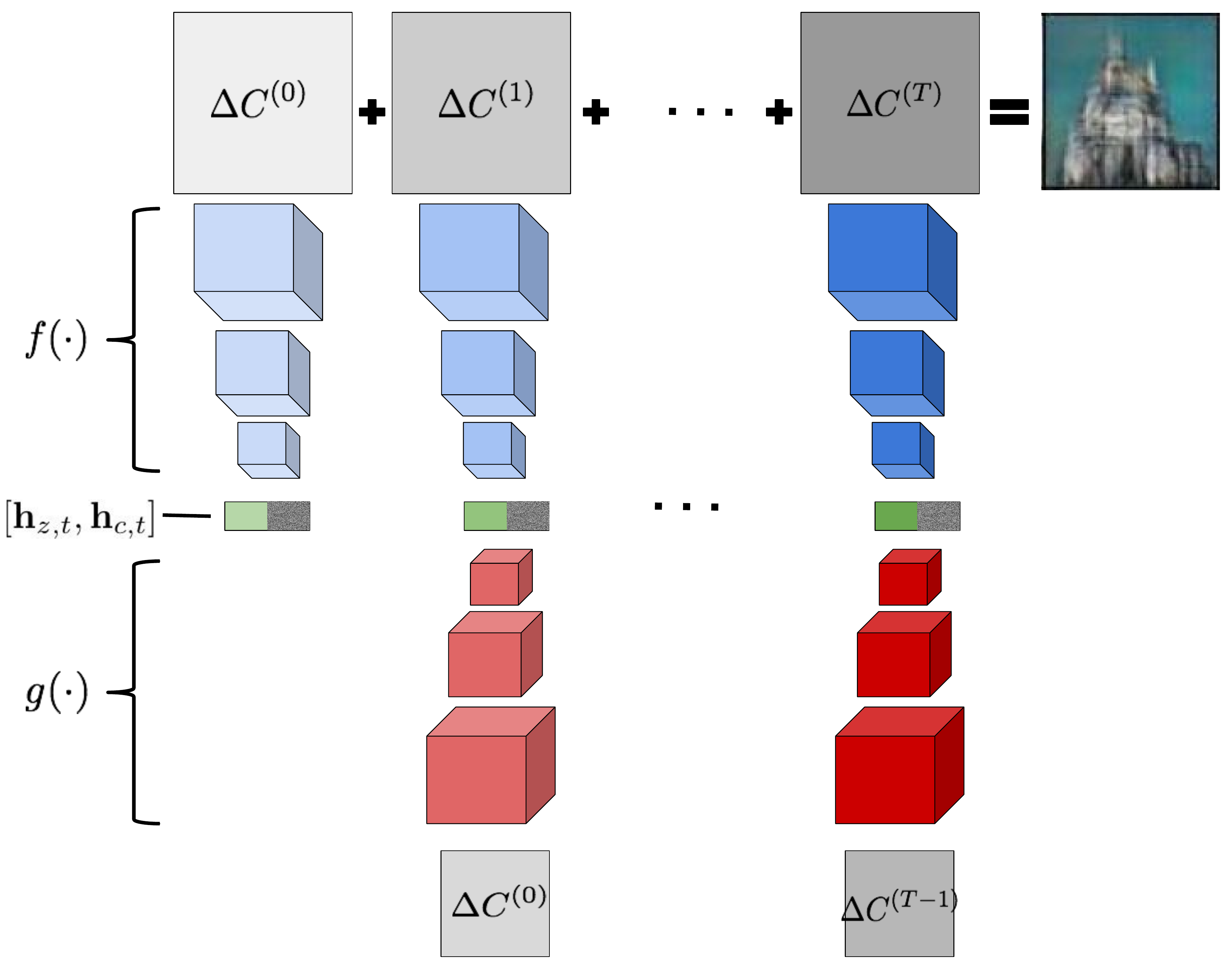}
    \vskip -0.05in
    \caption{Abstraction of Generative Recurrent Adversarial Networks. The function $f$
    serves as the decoder and the function $g$ serves as the encoder of GRAN.}
    \label{figs:GRAN_skeleton}
\vskip -0.05in
%\end{wrapfigure} 
\end{figure}

%Previously, we motivated why it is worth exploring 
%sequential modeling even for stationary data. 
We propose Generative Recurrent Adversarial Networks (GRAN), 
whose underlying structure is similar to other GANs.
The main difference between GRAN versus other generative adversarial models is that
the generator $G$ consists of a recurrent feedback loop that takes
a {\em sequence} of noise samples drawn from the prior distribution 
$\bz \sim p(\bz)$ and draws an ouput at multiple time steps
$\Delta C_1, \Delta C_2, \cdots, \Delta C_T$. 
Accumulating the updates at each time step yields 
the final sample drawn to the canvas $\mathcal{C}$. 
Figure~\ref{figs:GRAN_skeleton} delineates
the high-level abstraction of GRAN. 
%which has 
%similar spirits as the recurrent neural networks or a decoder part of DRAW,
%and it may also resemble other sequential models as well. 

At each time step, $t$, a sample $\bz$ from the prior distribution 
is passed to a function $f(\cdot)$ along with the hidden states $\mathbf{h}_{c,t}$. 
Where $\mathbf{h}_{c,t}$ represent the hidden state, or in other words, a
 current encoded status of the previous drawing $\Delta C_{t-1}$.
%More specifically, 
%$\mathbf{h}_{c,t}$ is a hidden state that is encoded by function $g(\cdot)$
%from the previous drawing $\Delta C_{t-1}$. 
Here, $\Delta C_t$ %is what is drawn to the canvas at time $t$ and it 
represents the output of the function $f(\cdot)$. (see Supp. Figure 11.)
%which is responsible for decoding the noise 
%sample $\bz_t$ and hidden state $\mathbf{h}_{c,t}$. 
Henceforth, the function $g(\cdot)$
can be seen as a way to mimic the inverse of the function $f(\cdot)$.

Ultimately, the function $f(\cdot)$ acts as a decoder that receives 
the input from the previous hidden state $\mathbf{h}_{c,t}$ and 
noise sample $\bz$, and function $g(\cdot)$ acts as an encoder
that provides a hidden representation of the output $\Delta C_{t-1}$ for time step $t$. 
One interesting aspect of GRAN is that the procedure of GRAN starts
with a decoder instead of an encoder. This is in contrast to most auto-encoder like models
such as VAE or DRAW, which start by encoding an image (see Figure~\ref{figs:GRAN_skeleton}). 
%This is due to the innate property of generative adversarial 
%models, which does not have to infer the posterior distribution of an input. 

In the following, we describe the procedure in more detail. 
We have an initial hidden state $\mathbf{h}_{c,0}$ that %where $\mathbf{h}_{c,0}$ 
is set as
a zero vector in the beginning. %Furthermore, $\mathbf{h}_{z,t}$ is 
%initialized as 
%\begin{align}
%    %\bz &\sim p(Z)\\
%    \mathbf{h}_{z,0} &= \tanh(W \bz + \mathbf{b}). \label{eqn:affine_z}
%\end{align}
We then compute the following for each time step $t=1\ldots T$:
\begin{align}
    \bz_t &\sim p(Z)\\
    \mathbf{h}_{c,t} &= g(\Delta C_{t-1})\\
    \mathbf{h}_{z,t} &= \tanh(W \bz_t + \mathbf{b}). \label{eqn:affine_z}\\
    \Delta C_t &= f([\mathbf{h}_{z,t}, \mathbf{h}_{c,t}]),
\end{align}
where $[\mathbf{h}_{z,t}, \mathbf{h}_{c,t}]$ denotes the concatenation of 
$\mathbf{h}_{z,t}$ and $\mathbf{h}_{c,t}$
\footnote{Note that we explore two scenarios of sampling $\bz$ in the experiments.
The first scenario is
where $\bz$ is sampled once in the beginning, then $\mathbf{h}_{z,t}=\mathbf{h}_{z}$
as a consequence. 
In whe other scenario, $\bz$ is sampled at every time step.}. 
Finally, we sum the generated images and apply the logistic function 
 in order to scale the final output to be in $(0,1)$:  
\begin{align}
    \mathcal{C} = \sigma(\sum^{T}_{t=1} \Delta C_t).
\end{align}
%$\mathcal{C} = \sigma(\sum^{T}_{t=1} \Delta C_t)$.
The reason for using $tanh(\cdot)$ in 
Equation~\ref{eqn:affine_z} is to rescale $\bz$ to $(-1,1)$. 
Hence, rescaling it to the same (bounded) domain as $\mathbf{h}_{c,t}$. %and $g(\cdot)$, which is also applied with a
%$tanh(\cdot)$ nonlinearity. %, which makes $\mathbf{h}_{c,t}$ in the same range. %of $(-1,1)$. 

\begin{figure*}[t]
    \includegraphics[width=2\columnwidth]{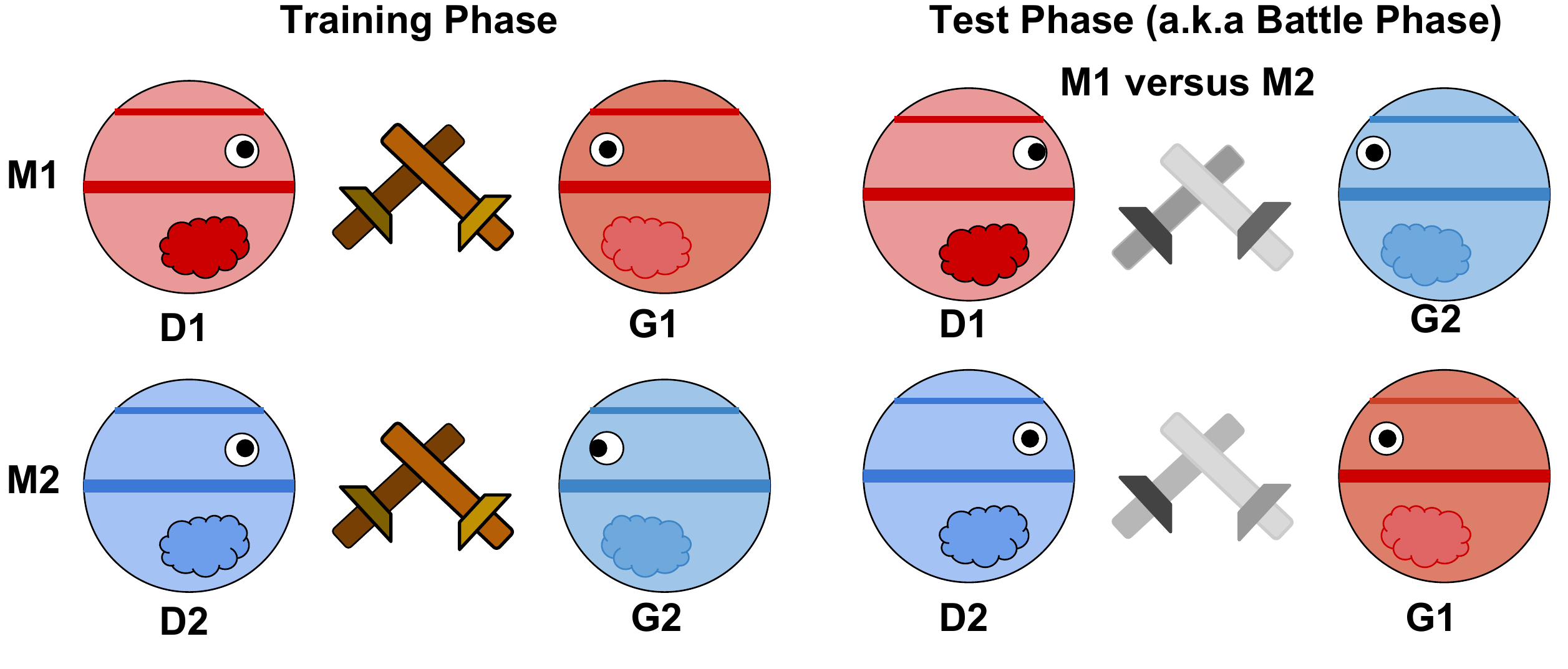}
    \vskip -0.05in
    \caption{Training Phase of Generative Adversarial Networks.}
    \label{figs:battle_train}
    \vskip -0.1in
\end{figure*} 

In general, one can declare the functions $f(\cdot)$ and $g(\cdot)$ to be any type of model. 
We used a variant of DCGAN \cite{Radford2015} in our experiments.
Supp. Figure 11 demonstrates the architecture of GRAN 
at time step $t$. The function $f(\cdot)$ starts with
one fully connected layer at the bottom and a deconvolutional
layers with fractional-stride convolution at rest of the upper layers.
This makes the images gradually upscale as we move up to higher layers.
Conversely, the function $g(\cdot)$ starts from convolutional layers
and the fully connected layer at the top. The two functions, 
$f(\cdot)$ and $g(\cdot)$, are symmetric copies of one another, 
as shown in Figure~\ref{figs:GRAN_skeleton}. 
The overall network is trained via backpropagation through the time.

%\begin{figure*}[t]
%    \begin{minipage}{0.5\textwidth}
%        \includegraphics[width=\columnwidth]{GAN_training.pdf}
%        \vskip -0.05in
%        \caption{Training Phase of Generative Adversarial Networks.}
%        \label{figs:battle_train}
%    \end{minipage}
%    \begin{minipage}{0.5\textwidth}
%        \includegraphics[width=\columnwidth]{GAN_test_phase.pdf}
%        \vskip -0.05in
%        \caption{Training Phase and Test Phase of Generative Adversarial Networks.}
%        \label{figs:battle_test}
%    \end{minipage}
%    \vskip -0.1in
%\end{figure*} 

\section{Model Evaluation: Battle between GANs}
\label{sec:model_sel}
A problem with generative adversarial models is
that there is no obvious way to evaluate them quantitatively.  
In the past, \cite{Goodfellow2014}  evaluated GANs by
looking at nearest-neighbours in the training data. 
LAPGAN was evaluated in the same way,
and in addition using human inspections \cite{Denton2015}.
For these, volunteers were asked to judge whether given
images are drawn from the dataset or generated by LAPGAN. 
In that case, the human acts as the discriminator, 
while the generator is a trained GAN.
The problem with this approach is that human inspectors can be subjective to high variance, 
which makes it necessary to average over a large number of these, and
the experimental setup is expensive and cumbersome. 
A third evaluation scheme, used recently by \cite{Radford2015} is based on classification 
performance. However, this approach is rather indirect and relies heavily on the
choice of  classifier. For example, in the work by Radford et al, they used nearest neighbor classifiers, 
which suffers from the problem that Euclidean distance is not a good dissimilarity measure 
for images. 

\begin{figure*}[t]
    \begin{minipage}{0.5\textwidth}
        \centering
        \vskip -0.025in
        \includegraphics[scale=0.9,width=\columnwidth]{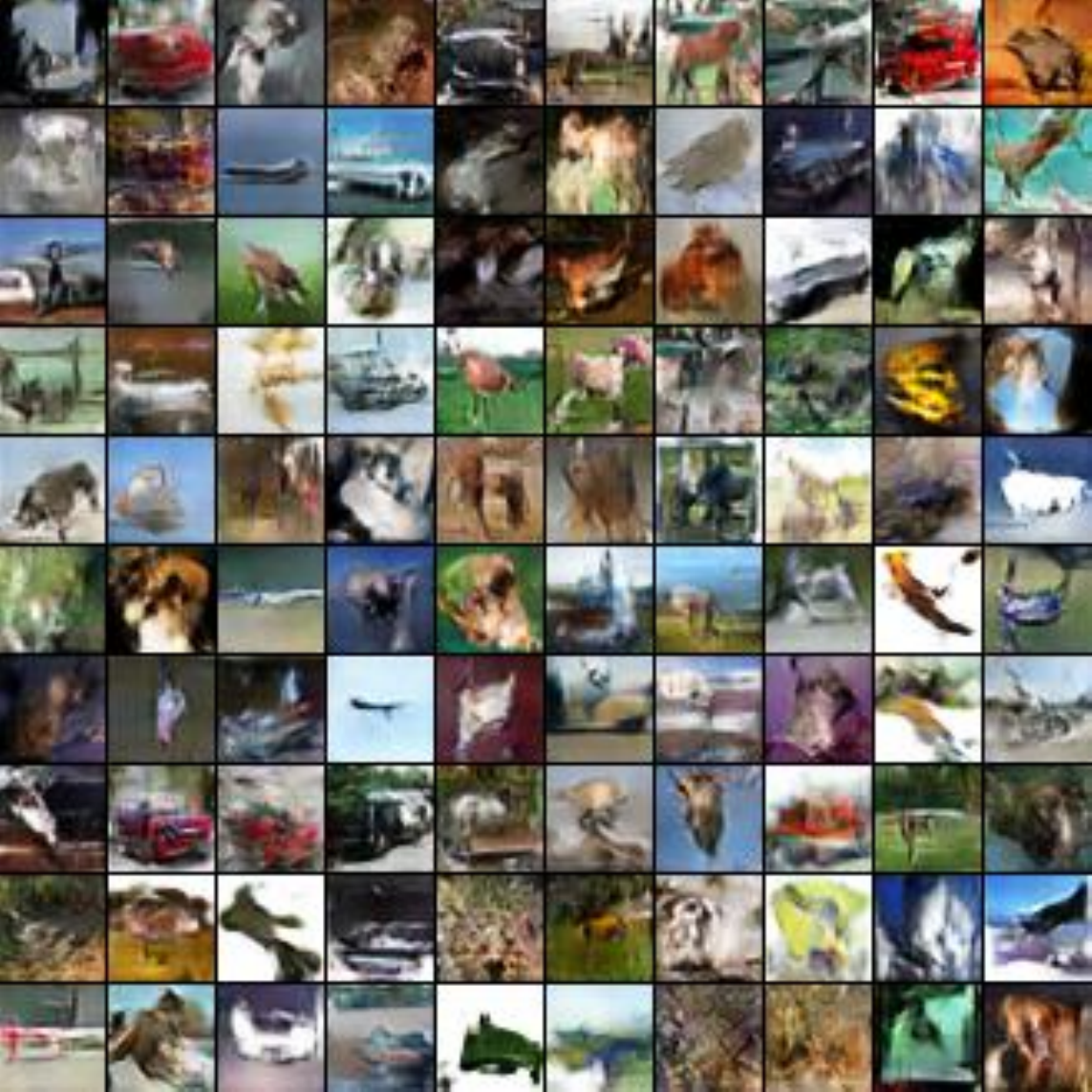}
        \vskip -0.05in
        \caption{Cifar10 samples generated by GRAN}
        \label{figs:cifar10_samples}
    \end{minipage}
    \begin{minipage}{0.5\textwidth}
        \centering
        \includegraphics[width=\columnwidth]{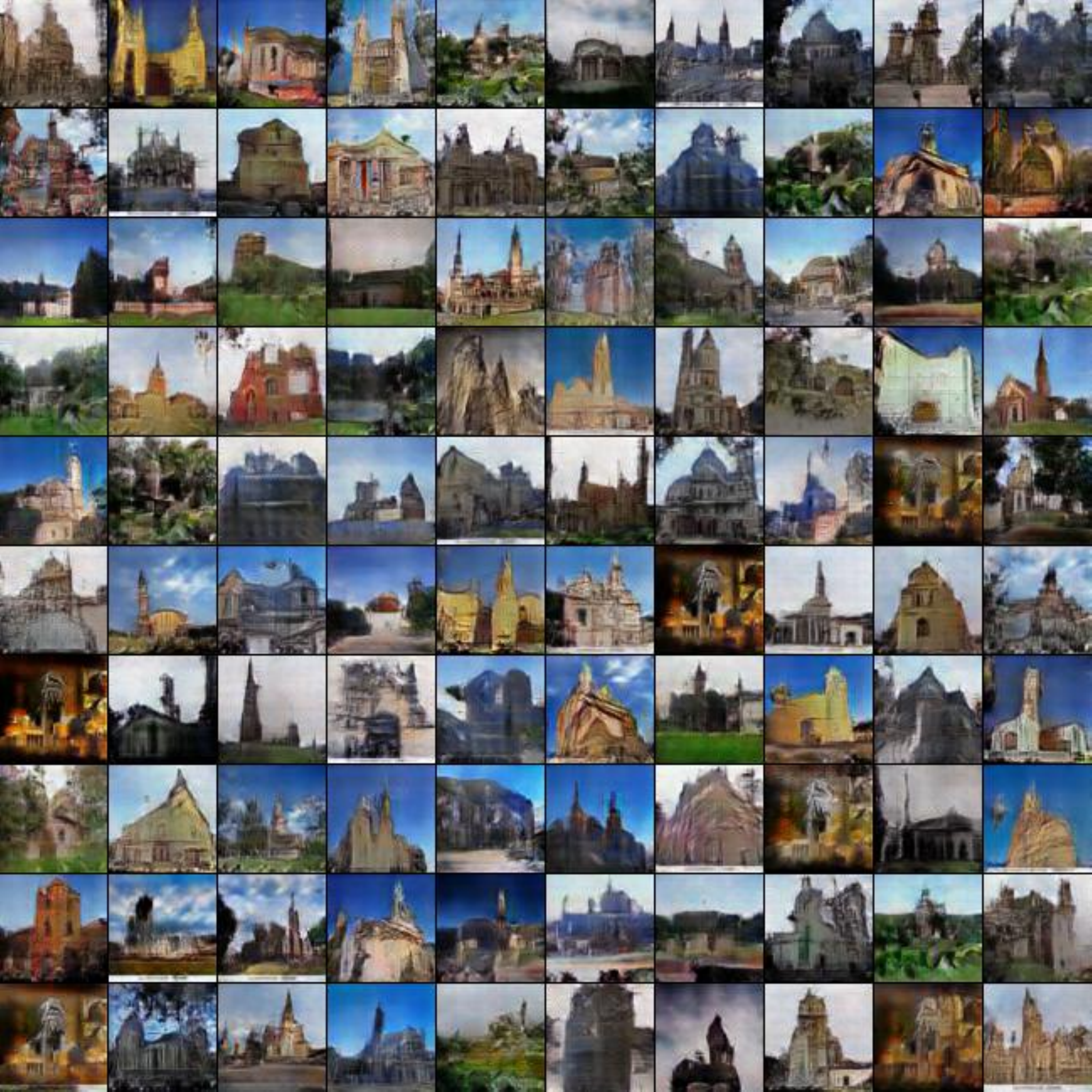}
        \vskip -0.05in
        \caption{LSUN samples generated by GRAN}
        \label{figs:lsun_samples}
    \end{minipage}
\vskip -0.1in
\end{figure*}

Here, we propose an alternative way to evaluate generative adversarial models.
Our approach is to directly compare two generative adversarial models 
by having them engage in a ``battle'' against each other. 
The naive intuition is that, since 
every generative adversarial models consists of a discriminator and a generator 
in pairs, we can exchange the pairs and have the models play the generative adversarial 
game against each other. 
Figure~\ref{figs:battle_train} illustrates this approach. 

The training and test stages are as follows.
Consider two generative adversarial models, $M_1$ and $M_2$.
Each model consists of a generator and a discriminator, 
$M_1 = \lbrace (G_1, D_1) \rbrace$ and $M_2 = \lbrace(G_2,D_2)\rbrace$.
%\begin{align}
%    M_1 = \lbrace (G_1, D_1) \rbrace \text{ and } 
%    M_2 = \lbrace(G_2,D_2)\rbrace.
%\end{align}
%During the training stage, both models are being trained to prepare them for the battle with one another. Thus, 
In the training phase, $G_1$ competes
with $D_1$ in order to be trained for the battle in the test phase.
Likewise for $G_2$ and $D_2$. In the test phase,
model $M_1$ plays against model $M_2$ by having $G_1$ try to fool $D_2$ 
and vice-versa.

\begin{table}[htp]
%\begin{wrapfigure}{l}{0.66\textwidth}
\vspace{-0.1cm}
\caption{Model Comparison Metric for GANs}
\label{tab:metric}
\begin{center}
\begin{small}
\begin{sc}
\begin{tabular}{ccc}
\hline
%\abovespace\belowspace
& $\bm{M_1}$ & $\bm{M_2}$   \\
\hline
%\abovespace
$\bm{M_1}$ & $D_1(G_1(\bz))$ , $D_1(\bx_{train})$ & $D_1(G_2(\bz))$ , $D_1(\bx_{test})$  \\
%\belowspace                                                      
$\bm{M_2}$ & $D_2(G_1(\bz))$ , $D_2(\bx_{test})$  & $D_2(G_2(\bz))$ , $D_2(\bx_{train})$ \\
\hline
\end{tabular}
\end{sc}
\end{small}
\end{center}
\vskip -0.15in
%\end{wrapfigure} 
\end{table}

Accordingly, we end up with the combinations shown in
Table~\ref{tab:metric}. Each entry in the table contains
two scores, one from discriminating training or test data points, and
the other from discriminating generated samples. At test time,
we can look at the following ratios between the discriminative scores of 
the two models: 
\begin{align}
    r_{test}   &\eqdef \frac{\epsilon \big(D_1(\bx_{test})\big)}{\epsilon\big(D_2(\bx_{test})\big)} \text{ and } 
    r_{sample} \eqdef \frac{\epsilon \big(D_1(G_2(\bz))  \big)}{\epsilon\big(D_2(G_1(\bz))  \big)},
\end{align}
where $\epsilon(\cdot)$ is the classification error rate, and $x_{test}$ is the predefined test set.
These ratios allow us to compare the model performances. 

The test ratio, $r_{test}$, tells us which model generalizes better
since it is based on discriminating the test data. Note that 
when the discriminator is overfitted to the training data, 
the generator will also be affected by this. 
This will increase the chance of producing biased samples towards 
the training data.

The sample ratio, $r_{sample}$, tells us 
which model can fool the other model more easily, since the discriminators are
classifying over the samples generated by their opponents. Strictly speaking,
as our goal is to generate good samples,
the sample ratio determines which model is better at generating good 
(``data like'') samples. 
%When the error rate of $D_1$ is lower than $D_2$, then the ratio will be less than 1 and when $D_2$ is lower than $D_1$, then the ratio will be greater than 1 as shown:

We suggest using the sample ratio to determine the winning model, and to use 
the test ratio to determine the validity of the outcome. %  as outlined below. 
The reason for using the latter is 
due to the occasional possibility of the sample ratio being biased, in which 
case the battle is not completely fair when the winner is solely determined by 
the sample ratio. 
%Our objective is to figure out which model is better at generating good 
%(“data like”) samples based on sample ratio ($r_sample$). 
%If we were to only consider the sample ratio in deciding the winner, then 
%there is a possibility for the sample ratio to be biased. 
%computing the sample ratio requires discriminators 
%from each model, and 
It is possible that one of the discriminators is biased towards the 
training data more so than the other (i.e. overfitted on the training data). 
%Consider as a counter example the case where the discriminator of $M_1$ only outputs
%false and the generator of $M_1$ is trained against the discriminator of $M_1$. 
%On the other hand, $M_2$ is a model that is trained based on generative 
%adversarial objective in Equation~\ref{eqn:gan_pac_obj}.
%Then, the error rate for $D_1$ on samples generated by $M_2$ will be zero.
%So, $M_1$ wins since the error rate of $M_1$ is lower than error rate of $M_2$. 
%However, $M_1$ should lose to $M_2$ since $M_2$ is obviously not a good model. 
%This problem arises because we have not accounted for the test ratio.  
In order to address this issue, our proposed evaluation metric qualifies the sample ratio to be judged by 
% by defining the winning model
the test ratio as follows: 
\begin{equation}
\text{winner} = 
    \begin{cases} 
        $M1$ & \text{if } r_{sample} < 1 \; \; \mathrm{and} \; \; r_{test}\simeq 1\\
        $M2$ & \text{if } r_{sample} > 1 \; \; \mathrm{and} \; \; r_{test}\simeq 1\\
        \text{Tie} & \text{otherwise } 
   \end{cases}
   \label{eqn:sample_ratio_metric}
\end{equation}
This imposes a condition where $r_{test} \simeq 1$, which 
assures that none of the discriminator is overfitted more than the other. 
If $r_{test} \neq 1$, then this implies that $r_{sample}$ is biased,
and thus, the sample ratio is no longer applicable.

%\begin{definition}
%    {\bf (Generative Adversarial Metric)}\\ Equation~\ref{eqn:sample_ratio_metric}
%    is a valid evaluation metric if the test ratio is approximately equal to 1, 
%     $r_{test}\simeq 1$ .
%\end{definition}
%
%The reason why there exist such a condition on generative adversarial metric (GAM) is to ensure that both discriminators have the similar performance. 
%, this means that discriminators of two models are about the same level with
%one and the other.
%This makes the decision based on sample ratio to be fair. Moreover, this condition should be satisfied since we want the two discriminators to perform roughly the same on the test data.
We call this evaluation measure Generative Adversarial Metric (GAM). 
GAM is not only able to compare generative adversarial models against each other, 
but also able to partially compare other models, such as the VAE or DRAW.
This is done by observing the error rate of GRAN's discriminators based on the samples 
of the other generative model as an evaluation criterion. For example, in our 
experiments
we report the error rates of the GRAN's discriminators with 
the samples of other generative models, i.e. $err(D(\bz))$ where $\bz$ are the samples of 
other generative models and $D(\cdot)$ is the discriminator of GRAN.

%For example, let the discriminator of GAN discriminate
%samples generated by VAE. Although, this may not be the best way to assess the two models,
%we tested and report the results of these comparisons in the experiments.

\section{Experiments}
In order to evaluate whether the extension of sequential generation enhances the 
perfomance, 
we assessed both quantitatively and qualitatively under three different image datasets. 
We conducted several empirical studies on GRAN under the model selection metrics 
discussed in Section~\ref{sec:model_sel}. 
%Additionally, we inspected generated samples them to evaluate the model qualitatively.
See Supplementary Materials %Appendix~\ref{sup:exp}
for full experimental details. 
%\begin{wrapfigure}{l}{0.66\textwidth}
\begin{table*}[t]
    \vskip -0.15in
    \begin{minipage}{0.66\textwidth}
        \caption{Model Evaluation on various data sets.}
        \label{tab:gran}
        \begin{center}
        {\small 
        %\begin{sc}
        \begin{tabular}{lcccc}
        \hline
        %\abovespace\belowspace
        Data set & Battler & r$_{test}$ & r$_{sample}$ & Winner\\
        \hline
        %\abovespace
        \multirow{2}*{MNIST }   & GRAN1 vs. GRAN3 & 0.79 & 1.75  & GRAN3 \\
                                & GRAN1 vs. GRAN5 & 0.95 & 1.19  & GRAN5 \\\hline
        \multirow{3}*{CIFAR10}  & GRAN1 vs. GRAN3 & 1.28 & 1.001 & GRAN3 \\
                                & GRAN1 vs. GRAN5 & 1.29 & 1.011 & GRAN5 \\
                                & GRAN3 vs. GRAN5 & 1.00 & 2.289 & GRAN5 \\\hline
        \multirow{3}*{LSUN}     & GRAN1 VS. GRAN3 & 0.95 & 13.68 & GRAN3 \\
                                & GRAN1 vs. GRAN5 & 0.99 & 13.97 & GRAN5 \\
        %\belowspace
                                & GRAN3 vs. GRAN5 & 0.99 & 2.38   & GRAN5 \\
        \hline
        \end{tabular}
        %\end{sc}
        }
        \end{center}
    \end{minipage}
    \begin{minipage}{0.33\textwidth}
    \vskip -0.15in
    \caption{Comparison between GRAN and non-adversarial models on MNIST.}
    \label{tab:battle_non_gans}
        \begin{center}
        {\small 
        %\begin{sc}
        \begin{tabular}{cc}
        \hline
        %\abovespace\belowspace
        Battler &  Error  \\
        \hline
        %\abovespace
        GRAN1 vs. DVAE &  0.058 \\
        GRAN3 vs. DVAE &  0.01  \\
        GRAN1 vs. DRAW &  0.347 \\
        %\belowspace
        GRAN3 vs. DRAW &  0.106 \\
        \hline
        \end{tabular}
        %\end{sc}
        } \end{center}
    \end{minipage}
    \vskip -0.2in
%\end{wrapfigure}
\end{table*} 

In the following, we analyze the results by answering a set of questions 
on our experiments.

\textit{Q: How does GRAN perform?}

The performance of GRAN is presented in Table~\ref{tab:gran}.
We focused on comparing GRANs with 1, 3 and 5 time steps. 
%which were denoted as GRAN1, GRAN3 and GRAN5. 
For all three datasets,
GRAN3 and GRAN5 outperformed GRAN1 as shown in Table~\ref{tab:gran}. 
Moreover, we present samples from GRAN for MNIST, cifar10 and LSUN in  
Figure~\ref{figs:cifar10_samples}, Figure~\ref{figs:lsun_samples}, and Supp. Figure 23.
%Most of the MNIST samples are shown in Supplementary Materials. 
Figure 23 and Figure~\ref{figs:cifar10_samples} 
appear to be discernible and reasonably classifiable by humans. 
Additionally, the LSUN samples from Figure~\ref{figs:lsun_samples} seem to cover 
 variety of church buildings and contain fine detailed textures.
The ``image statistics'' of two real image datasets are 
embedded into both types of sample.

%\begin{wrapfigure}{r}{0.33\textwidth}
%%\begin{table*}[t]
%    \begin{minipage}{0.33\textwidth}
%    \caption{Comparison between GRAN and non-adversarial models on MNIST.}
%    \label{tab:battle_non_gans}
%    \vskip 0.15in
%    \begin{center}
%    \begin{small}
%    \begin{sc}
%    \begin{tabular}{cc}
%    \hline
%    %\abovespace\belowspace
%    Battler &  Error  \\
%    \hline
%    %\abovespace
%    GRAN1 vs. DVAE &  0.058 \\
%    GRAN3 vs. DVAE &  0.01  \\
%    GRAN1 vs. DRAW &  0.347 \\
%    %\belowspace
%    GRAN3 vs. DRAW &  0.106 \\
%    \hline
%    \end{tabular}
%    \end{sc}
%    \end{small}
%    \end{center}
%    \end{minipage}
%    \vskip -0.2in
%\end{wrapfigure}
%%\end{table*} 
\textit{Q: How do GRAN and other GAN type of models perform compared to 
non generative adversarial models?}

We compared our model to other generative models such 
as denoising VAE (DVAE) \cite{Im2015} and DRAW on the MNIST dataset.
Although this may not be the best way to assess the two models, 
since the generator of GRAN is not used,
%since the generator of GRAN does get assessed explicitly, 
Table~\ref{tab:battle_non_gans} presents the results 
of applying GAM as described at the end of Section~\ref{sec:model_sel}.
The error rates were all below 50\%, and especially low for DVAE's samples. 
Surprisingly, even though samples from DRAW look very nice, the error rate 
on their samples were also quite low with GRAN3. 
This illustrates that the discriminators of generative adversarial models are 
good at discriminating the samples generated by DVAE and DRAW.
Our hypothesis is that the samples look nicer due to the smoothing effect 
of having a mean squared error in their objective, but they do not 
capture all relevant aspects of the statistics of real handwritten images. 

%%\textit{Q: How do GRAN's samples look? }
%
%We present samples from GRAN for MNIST, cifar10 and LSUN in  
%Supp. Figure~\ref{figs:mnist_samples}, Figure~\ref{figs:cifar10_samples} and Figure~\ref{figs:lsun_samples}.
%Most of the MNIST and cifar10 samples shown in
%Supp. Figure~\ref{figs:mnist_samples} and Figure~\ref{figs:cifar10_samples} 
%appear to be discernible and reasonably classifiable by humans. 
%Similarly, the LSUN samples from Figure~\ref{figs:lsun_samples} seem to cover 
% variety of church buildings and contain fine detailed textures.
%The ``image statistics'' of two real image datasets are 
%embedded into both types of sample.
%%One concern is that although none of the samples are identical to each other, but
%%few of the samples look close to one and the other. This collapsing samples 
%%is genetic problem with generative adversarial models as mentioned 
%%in \cite{Radford2015}. We notice that samples gradually
%%become similar to each other when the model is trained for too long. Hence, one must do early
%%stopping before samples collapse.
\begin{figure}[htp]
    \vskip -0.1in
    \begin{minipage}{0.49\textwidth}
        \centerline{\includegraphics[width=\columnwidth]{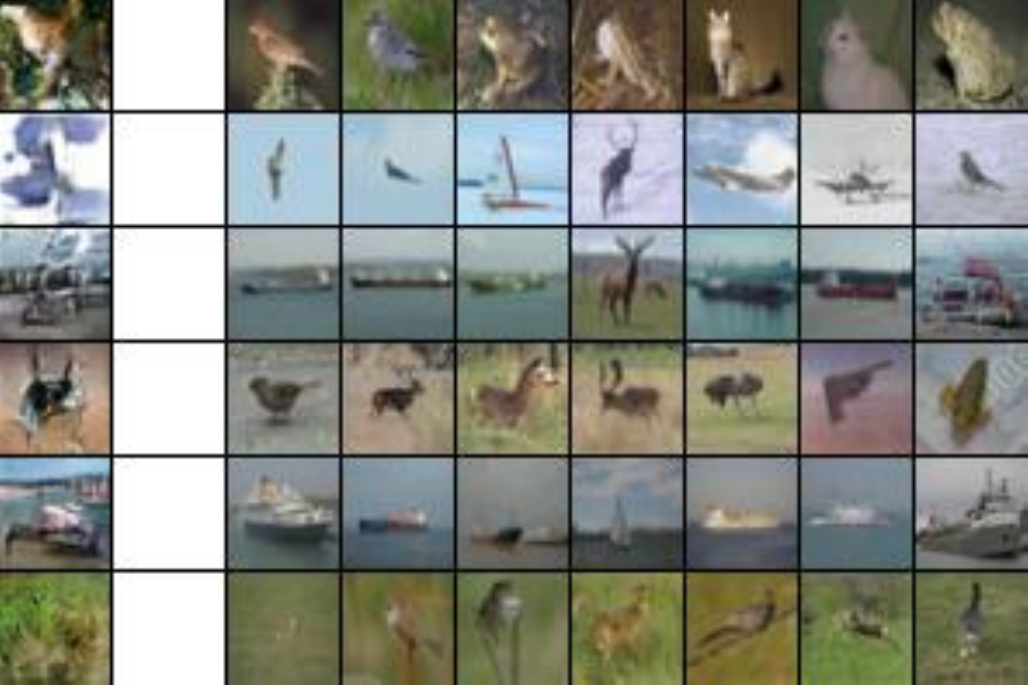}}
        \vskip -0.05in
        \caption{Nearest Neighbour training examples for cifar10 samples.}
        \label{figs:cifar10_NN_samples_main} 
    \end{minipage}
    \begin{minipage}{0.49\textwidth}
        \centerline{\includegraphics[width=\columnwidth]{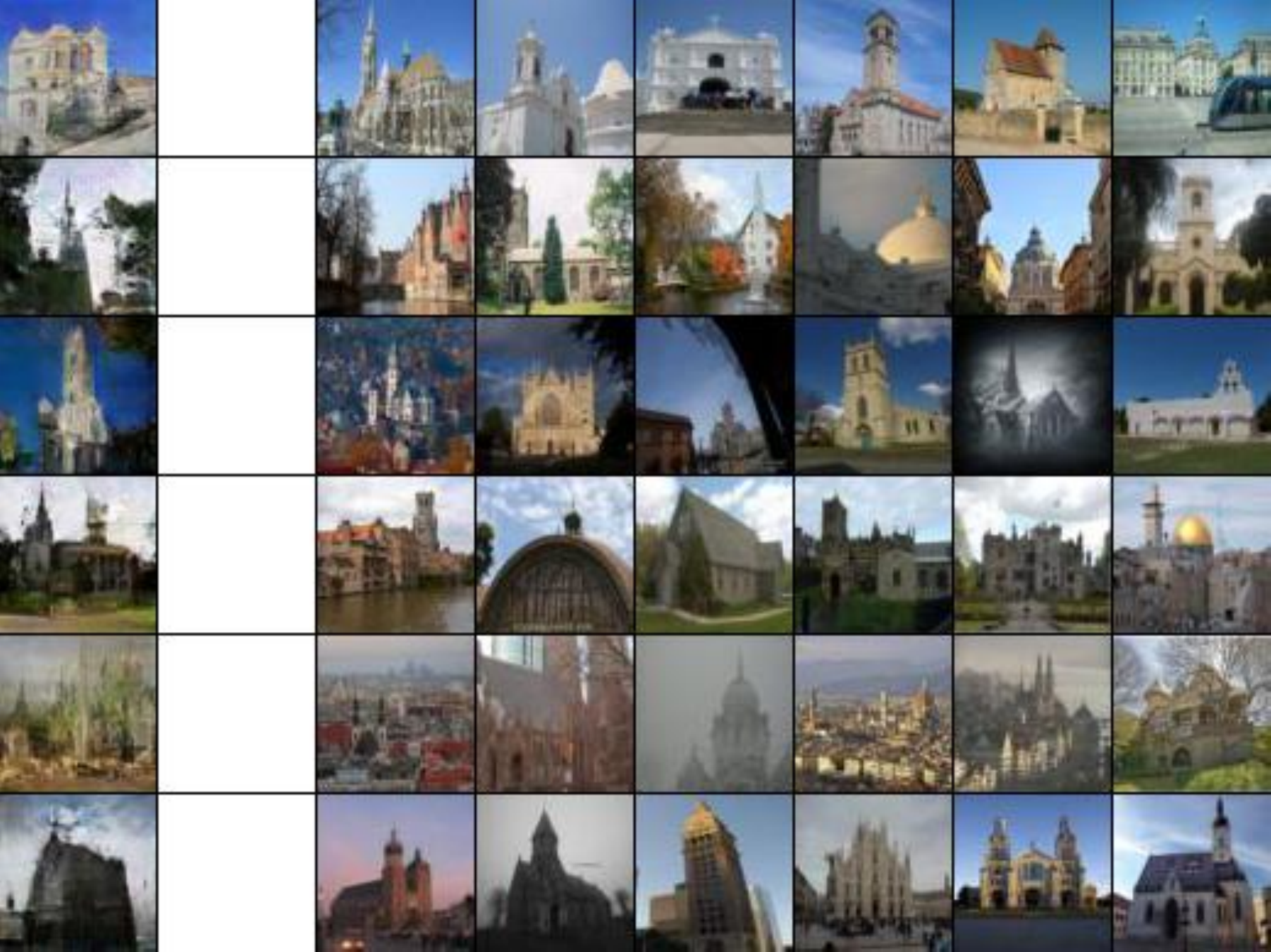}}
        \vskip -0.05in
        \caption{Nearest Neighbour training examples for lsun samples using GRAN3.}
        \label{figs:lsun_NN_samples_main}
    \end{minipage}
\end{figure}
\textit{Q: Does GRAN overfit to the training data?}

%The most naive way would be to iterate over all training data to see whether the samples look the same as training data. 
Since it is infeasible to naively examine the training data for similar looking images as GRAN's output, it is common (albeit somewhat 
questionable) to look at $k$-nearest neighbors to do a basic sanity check.
%Although the Euclidean distance in the image space is hopeless, 
As shown in Figure~\ref{figs:cifar10_NN_samples_main} and \ref{figs:lsun_NN_samples_main},
%Supp. Figure 16, and Supp. Figure 17 and 18, 
one does not find any replicates of training data cases.

Empirically speaking, we did notice that GRAN tends to generate samples by
interpolating between the training data. For example, Supp. Figure 26
illustrates that the church buildings consist of similar structure of
the entrance but the overall structure of the church has a different shape. Based on
such examples, we hypothesize that the overfitting for GRAN 
in the worst case may imply that the model learns to interpolate 
sensibly between training examples. 
This is not the typical way of the term overfitting is used for 
generative models, which usually refers to memorizing the data.
In fact, in adversarial training in general, the objective function is not based 
on mean squared error of the pixels which makes it not obvious 
how to memorize the training samples.
However, this could mean that it is difficult for these models to generate 
images that are interpolated from training data.

%\begin{figure}[htp]
%    \centerline{\includegraphics[width=\columnwidth]{cifar10_gran_nn_ts3_2.pdf}}
%    \caption{Nearest Neighbour training examples for cifar10 samples.}
%    \label{figs:cifar10_NN_samples}
%\vskip -0.2in
%\end{figure} 
\begin{figure}[htp]
    \begin{minipage}{0.49\textwidth}
        \begin{center}
        \includegraphics[width=\columnwidth]{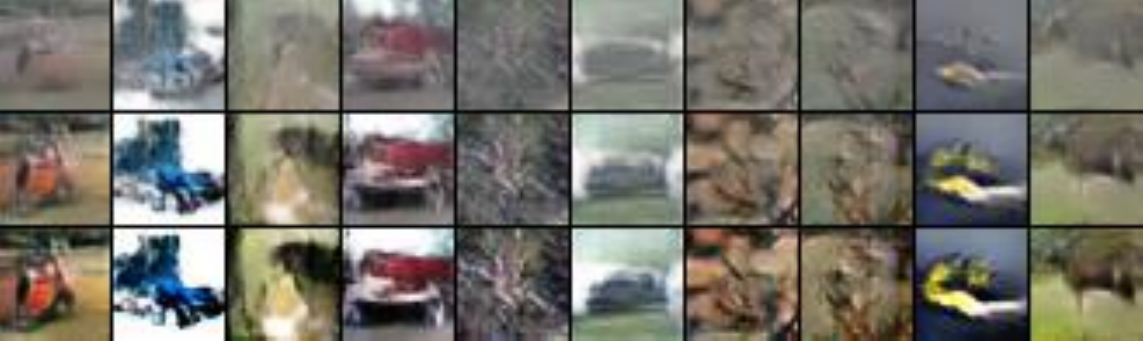}
        \vskip -0.05in
        \caption{Drawing at different time steps on cifar10 samples.}
        \label{figs:cifar10_seq_3step}
        \end{center}
    \end{minipage}
    \begin{minipage}{0.49\textwidth}
        \begin{center}
        \includegraphics[width=\columnwidth]{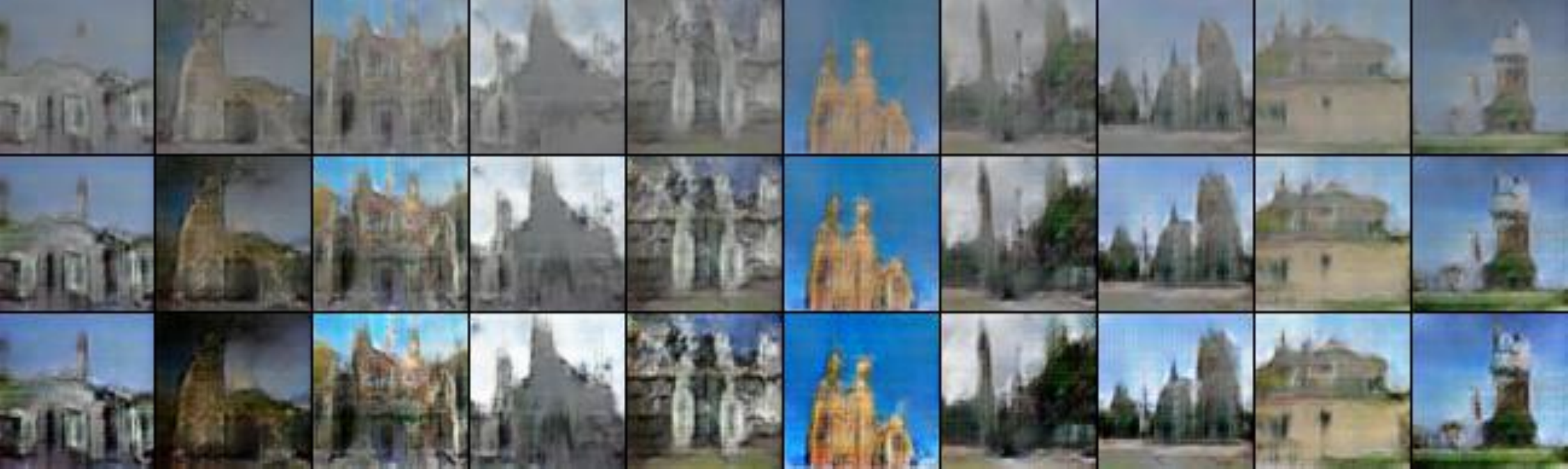}
        \vskip -0.05in
        \caption{Drawing at different time steps on lsun samples.}
        \label{figs:lsun_seq_3step}
        \end{center}
    \end{minipage}
    \vskip -0.1in
\end{figure} 

\begin{figure*}[t]
    \begin{minipage}{0.49\textwidth}
        \includegraphics[scale=0.9,width=\columnwidth]{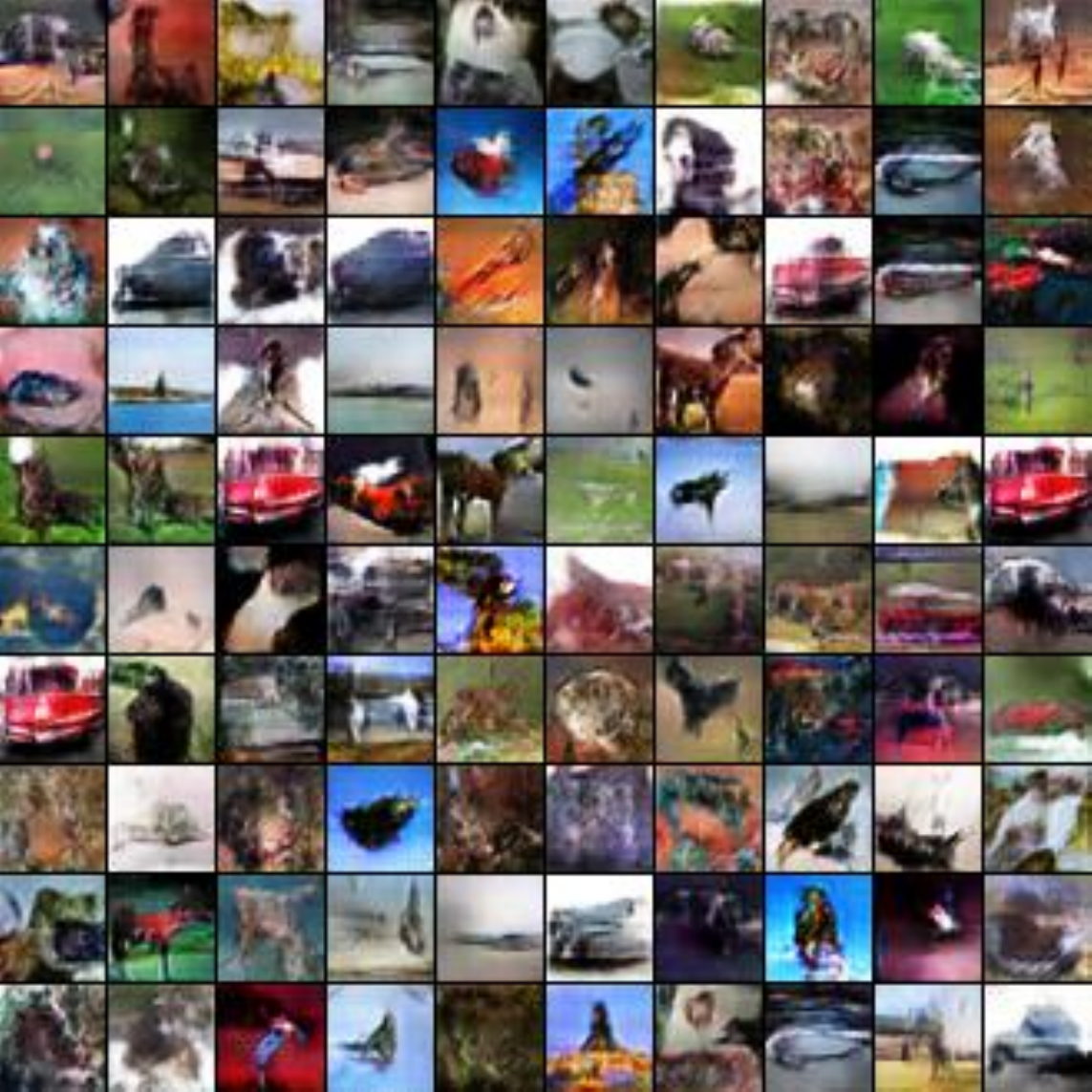}
        \vskip -0.05in
        \caption{Cifar10 samples generated by GRAN with injecting different noises at every time step}
        \label{figs:cifar10_samples_diff_Zs}
    \end{minipage}
    \begin{minipage}{0.49\textwidth}
        \includegraphics[width=\columnwidth]{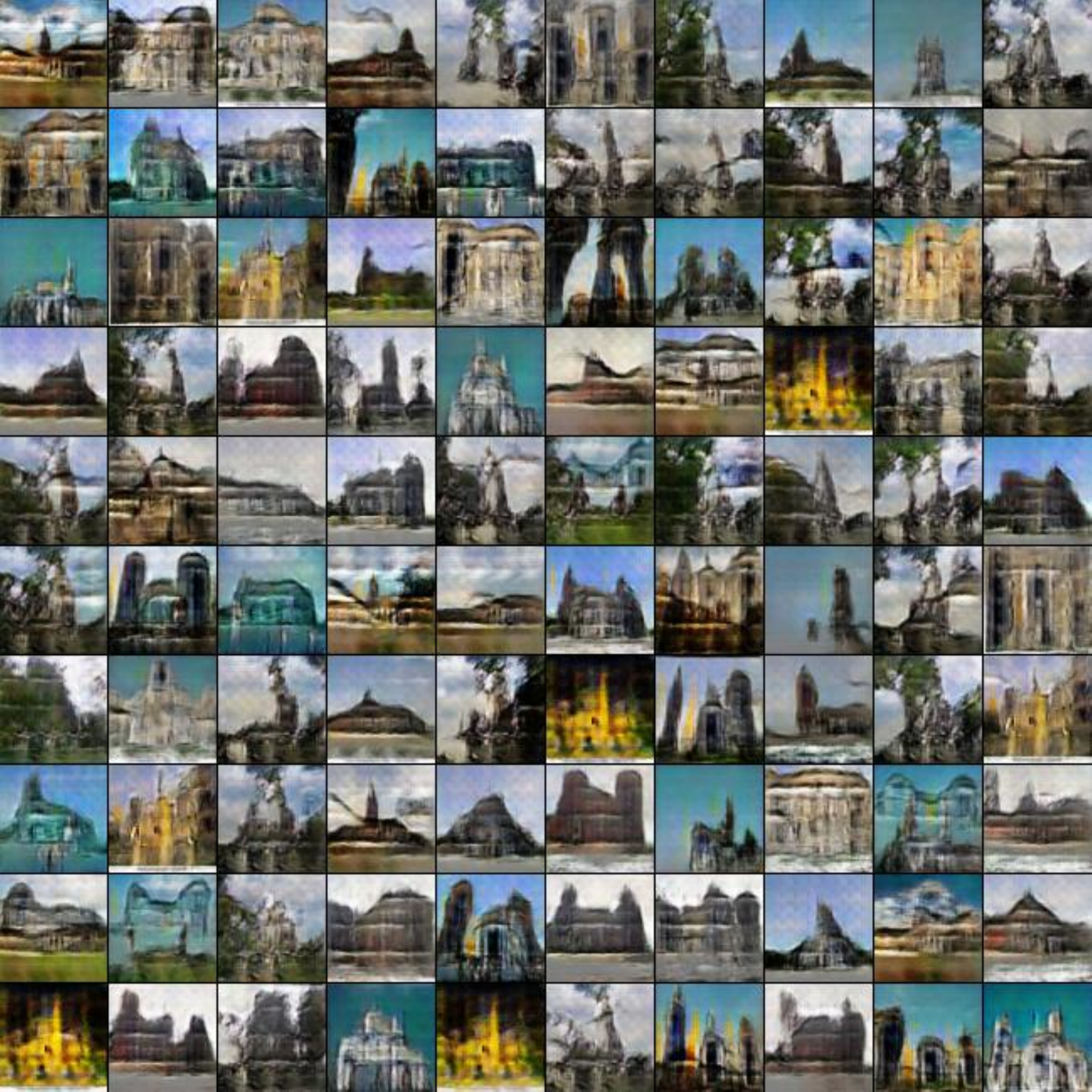}
        \vskip -0.05in
        \caption{LSUN samples generated by GRAN with injecting different noises at every time step}
        \label{figs:lsun_samples_diff_Zs}
    \end{minipage}
    \vskip -0.1in
\end{figure*}

{\em Q: What do the samples look during the intermediate time steps?}

Figure~\ref{figs:cifar10_seq_3step} and \ref{figs:lsun_seq_3step} present the 
intermediate samples when the total number of steps is $3$.
From the figures, we can observe the gradual development of the samples over time.
The common observation from the intermediate samples is that images become
more fine-grained and introduce details missing from the previous time step image.
Intermediate samples for models with a total number of time steps of $5$ can 
be found in the supplementary materials as well.
This behaviour is somewhat similar to \cite{Denton2015}, as one might expect 
(although filling-in of color details suggest that the process is more complex 
than a simple coarse-to-fine generation). 
Note that this behaviour is not enforced in our case, since we use an identical 
architecture at every time step. 

Next, we tested generating $7$ and $9$-step samples using GRAN5
which is trained with $5$- steps. 
These samples brighter compare 
to GRAN3 \& GRAN5 (See Supp. Figure 34 and 35).
But the quality of the samples look more or less visually 
similar.
When we evaluated them with GAM, the winner of the battle 
was GRAN7 as shown below:
%GRAN5 vs GRAN7 : r_test = 1.01933 and r_test = 0.947608.
%GRAN5 vs GRAN9 : r_test = 1.0908 and r_test = 1.07193.
%GRAN7 vs GRAN9 : r_test = 1.07011 and r_test = 0.958904.
\begin{table}[htp]
    \centering
    \caption{Model Evaluation on various data sets.}
    \label{tab:gran}
    {\small 
    \begin{tabular}{lccc}
    \hline
    %\abovespace\belowspace
    Data set & Battler & r$_{test}$ & r$_{sample}$ \\
    \hline
    %\abovespace
    \multirow{3}*{CIFAR10} & GRAN5 vs. GRAN7 & 1.01 & 0.94 \\
                           & GRAN5 vs. GRAN9 & 1.09 & 1.07 \\
                           & GRAN7 vs. GRAN9 & 1.07 & 0.95 \\\hline
    \end{tabular}
    }
\end{table}

\textit{Q: What happens when we use a different noises for each step?}

We sampled a noise vector $\mathbf{z}\sim p(Z)$ and used the same noise for 
every time step. This is because $\mathbf{z}$ acts as a reference frame in 
\cite{gatys2015neural} 
as shown in Figure~\ref{fig:drawartify}.
On the other hand, the role of the sample in DRAW is to inject noise at each step,
$\mathbf{z}_1, \mathbf{z}_2, \cdots \mathbf{z}_T,\sim p(Z)$, as prescribed by 
the variational auto-encoding framework. 
We also experimented with both sampling $\mathbf{z}$ once in the beginning
versus sampling $\mathbf{z}_i$ at each time step. Here we describe
the advantages and disadvantages to these two approaches.

The samples of cifar10 and LSUN generated by 
injecting different noises at each time step
are shown in Figure~\ref{figs:cifar10_samples_diff_Zs} and Figure~\ref{figs:lsun_samples_diff_Zs}.
Note that Figure~\ref{figs:cifar10_samples} and Figure~\ref{figs:lsun_samples}
were the output samples when injected using the same noise.
The samples appear to be discernible and reasonably classifiable by humans as well.
However, we observe a few samples that look very alike to one other.
During the experiments, we found that when using different noise, 
it requires more effort to find a set of hyper-parameters that 
produce good samples. Furthermore, the samples tend to collapse when training for a long time. 
Hence, we had to carefully select the total number of iterations. 
This illustrates that the training became much more difficult and 
it provokes GRAN to ``cheat'' by putting a lot of probability mass on samples that
the discriminator cannot classify, which produce samples that look 
very similar to each other.

On the other hand, when we look at the intermmediate time steps of samples
generated using multiple noises, we find that there are more pronounced changes 
within each time step as demonstrated in Supp. Figure 29
and Supp. Figure 30. For example, the colour of the train 
in Supp. Figure 29 changes, and a partial church is drawn in Supp. Figure 30. 
%This illustrates that adding different noise increases the model capability on generating much dynamical images.

%{\em Q: How do the samples look like during the intermediate time steps $t$?}
%
%Figure~\ref{figs:mnist_seq}, Figure~\ref{figs:cifar10_seq}, and Figure~\ref{figs:lsun_seq}
%presents the intermediate steps of samples when time step is 3.
%From the figures, we can observe the gradual development of the samples over time.
%The common notice from the intermediate transition is that the samples become
%finer and recovers missing details from the previous time step image.
%Intermediate samples for time step equal to 5 can be found in the supplementary
%materials.

%-------------------------------------------------------------------------
\section{Conclusion}
We proposed a new generative model based on adversarial training of a
recurrent neural network inspired by \cite{gatys2015neural}. 
We showed the conditions under which the model performs well 
and also showed that it can produce higher quality visual samples than an 
equivalent single-step model. We also introduced a new metric for 
comparing adversarial networks quantitatively and presented that the recurrent 
generative model yields a superior performance over existing state-of-the-art 
generative models under this metric.

%-------------------------------------------------------------------------
{\small
\clearpage
\newpage
\bibliographystyle{ieee}
\bibliography{gran}

\begin{thebibliography}{10}\itemsep=-1pt

\bibitem{Denton2015}
E.~Denton, S.~Chintala, A.~Szlam, and R.~Fergus.
\newblock Deep generative image models using a laplacian pyramid of adversarial
  networks.
\newblock In {\em Proceedings of the Neural Information Processing Systems
  (NIPS)}, 2015.

\bibitem{gatys2015neural}
L.~A. Gatys, A.~S. Ecker, and M.~Bethge.
\newblock A neural algorithm of artistic style.
\newblock {\em arXiv preprint arXiv:1508.06576}, 2015.

\bibitem{Gauthier2015}
J.~Gauthier.
\newblock Conditional generative adversarial nets for convolutional face
  generation.
\newblock In {\em Class Project for Stanford CS231N: Convolutional Neural
  Networks for Visual Recognition, Winter semester 2014}, 2014.

\bibitem{Goodfellow2014}
I.~J. Goodfellow, J.~Pouget-Abadie, M.~Mirza, B.~Xu, D.~Warde-Farley,
  S.~Ozair†, A.~Courville, and Y.~Bengio.
\newblock Generative adversarial nets.
\newblock In {\em Proceedings of the Neural Information Processing Systems
  (NIPS)}, 2014.

\bibitem{Gregor2015}
K.~Gregor, I.~Danihelka, A.~Graves, D.~J. Rezende, and D.~Wierstra.
\newblock Draw: A recurrent neural network for image generation.
\newblock In {\em Proceedings of the International Conference on Machine
  Learning (ICML)}, 2015.

\bibitem{hinton2006}
G.~E. Hinton, S.~Osindero, and Y.-W. Teh.
\newblock A fast learning algorithm for deep belief nets.
\newblock {\em Neural Computation}, 18(7):1527--1554, 2006.

\bibitem{Hyvrinen:2009:NIS:1572513}
A.~Hyvrinen, J.~Hurri, and P.~O. Hoyer.
\newblock {\em Natural Image Statistics: A Probabilistic Approach to Early
  Computational Vision.}
\newblock Springer Publishing Company, Incorporated, 1st edition, 2009.

\bibitem{Im2015}
D.~J. Im, S.~Ahn, R.~Memisevic, and Y.~Bengio.
\newblock Denoising criterion for variational auto-encoding framework.
\newblock In {\em arXiv preprint arXiv:1511.06406}, 2015.

\bibitem{Ioffe2015}
S.~Ioffe and C.~Szegedy.
\newblock Batch normalization: Accelerating deep network training by reducing
  internal covariate shift.
\newblock In {\em http://arxiv.org/pdf/1502.03167.pdf}, 2015.

\bibitem{Kingma2015}
D.~Kingma and J.~Ba.
\newblock Adam: A method for stochastic optimization.
\newblock In {\em Proceedings of the International Conference on Learning
  Representations (ICLR)}, 2014.

\bibitem{Kingma2014vae}
D.~P. Kingma and M.~Welling.
\newblock Auto-encoding varational bayes.
\newblock In {\em Proceedings of the Neural Information Processing Systems
  (NIPS)}, 2014.

\bibitem{Maas2013}
A.~L. Maas, A.~Y. Hannun, and A.~Y. Ng.
\newblock Rectifier nonlinearities improve neural network acoustic models.
\newblock In {\em Proceedings of the International Conference on Machine
  Learning (ICML)}, 2013.

\bibitem{Mirza2014}
M.~Mirza and S.~Osindero.
\newblock Conditional generative adversarial nets.
\newblock In {\em Proceedings of the Neural Information Processing Systems Deep
  learning Workshop(NIPS)}, 2014.

\bibitem{Nash1951}
J.~Nash.
\newblock Non-cooperative games.
\newblock {\em The Annals of Mathematics}, 54(2):286--295, 1951.

\bibitem{Radford2015}
A.~Radford, L.~Metz, and S.~Chintala.
\newblock Unsupervised representation learning with deep convolutional
  generative adversarial networks.
\newblock In {\em Proceedings of the International Conference on Learning
  Representations (ICLR)}, 2015.

\bibitem{ranzato2013modeling}
M.~Ranzato, V.~Mnih, J.~M. Susskind, and G.~E. Hinton.
\newblock Modeling natural images using gated mrfs.
\newblock {\em Pattern Analysis and Machine Intelligence, IEEE Transactions
  on}, 35(9):2206--2222, 2013.

\bibitem{Springenberg2014}
J.~T. Springenberg, A.~Dosovitskiy, T.~Brox, and M.~Riedmiller.
\newblock Striving for simplicity: The all convolutional net.
\newblock In {\em http://arxiv.org/abs/1412.6806}, 2014.

\bibitem{Yu2015}
F.~Yu, Y.~Zhang, S.~Song, A.~Seff, and J.~Xiao.
\newblock Lsun: Construction of a large-scale image dataset using deep learning
  with humans in the loop.
\newblock In {\em arXiv:1506.03365 [cs.CV] 10 Jun 2015}, 2015.

\bibitem{Zeiler2011}
M.~D. Zeiler, G.~W. Taylor, and R.~Fergus.
\newblock Adaptive deconvolutional networks for mid and high level feature
  learning.
\newblock In {\em International Conference on Computer Visio}, 2011.

\end{thebibliography}
}
\clearpage
\section*{Supplementary Materials}
\subsection*{Additional Notes on Convolutional Transpose}
\label{sec:supp_ct}
\begin{figure}[htp]
    \begin{minipage}{0.23\textwidth}
        \includegraphics[width=\columnwidth]{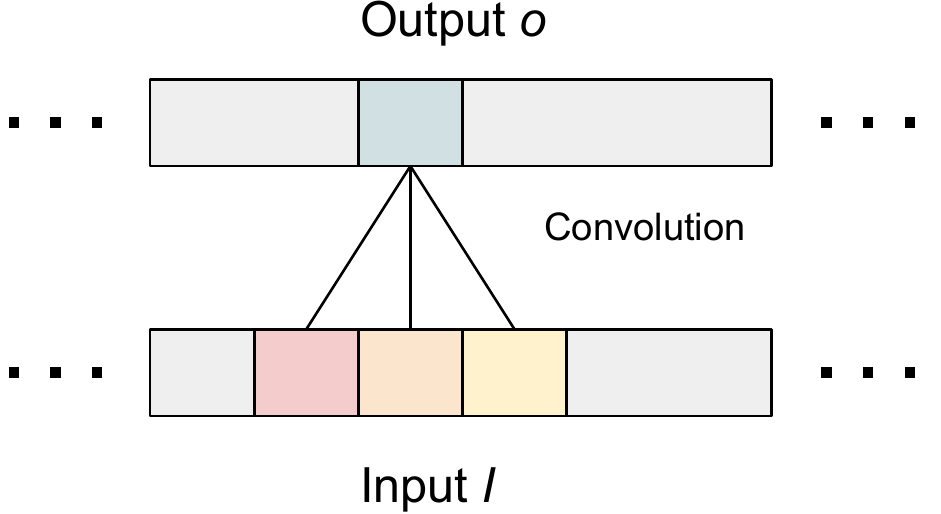}
        \caption{Applying convolution at index $j$.}
        \label{figs:convOp1}
    \end{minipage}
    \begin{minipage}{0.24\textwidth}
        \includegraphics[width=\columnwidth]{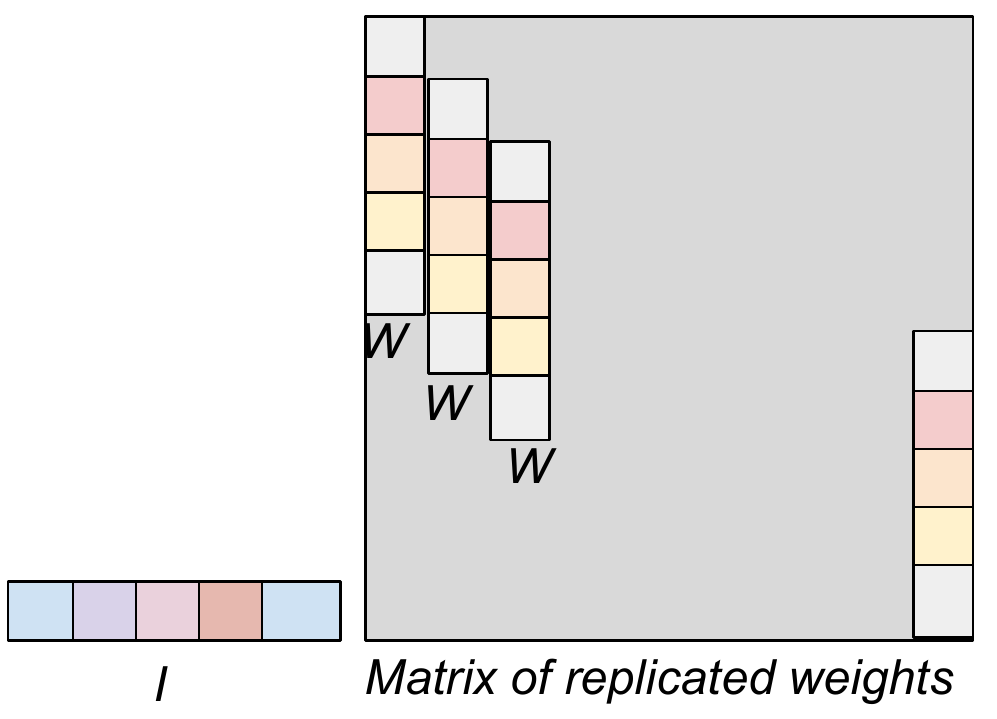}
        \caption{Convolution operation as a matrix operation.}
        \label{figs:convOp2}
    \end{minipage}
\vskip -0.2in
\end{figure}

In the following, we describe the convolutional transpose procedure in detail. 
For simplicity, let us consider the case of a 1-dimensional convolutional 
operation with one kernel and stride of 1: 
\begin{align}
    o = i * W,
    \label{eqn:convOp}
\end{align}
where $i$ is an input, $o$ is an output, and $*$ is the convolutiol operator.
Figure~\ref{figs:convOp1} and Figure~\ref{figs:convOp2} show an 
illustration of the 1D convolution.

Figure~\ref{figs:convOp1} presents the naive visualization of 
convolution over the input
centered at index $j$, and Figure~\ref{figs:convOp2} presents the 
convolution operation in terms of matrix operation.
The latter figure will be useful for understanding the convolutional transpose.

\begin{figure}[htp]
    \begin{minipage}{0.23\textwidth}
        \includegraphics[width=\columnwidth]{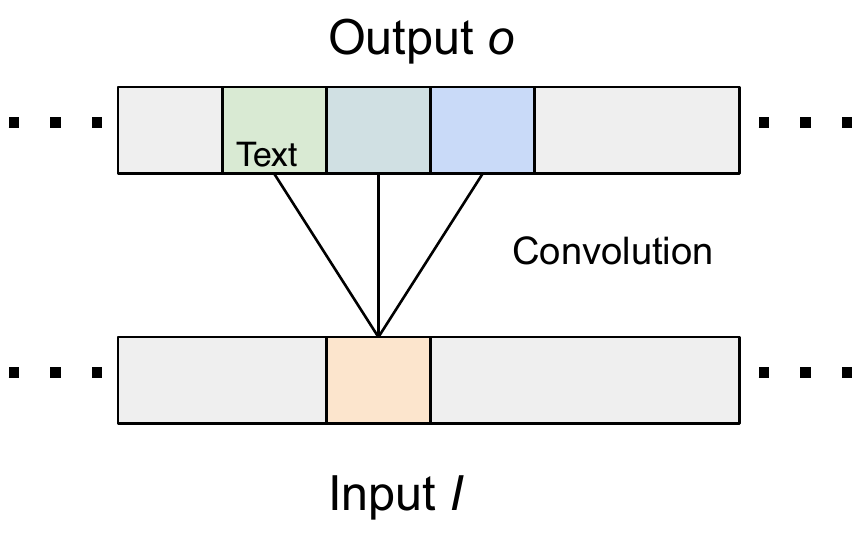}
        \caption{The gradient of convolution at index $k$.}
        \label{fig:gconvOp1}
    \end{minipage}
    \begin{minipage}{0.24\textwidth}
        \includegraphics[width=\columnwidth]{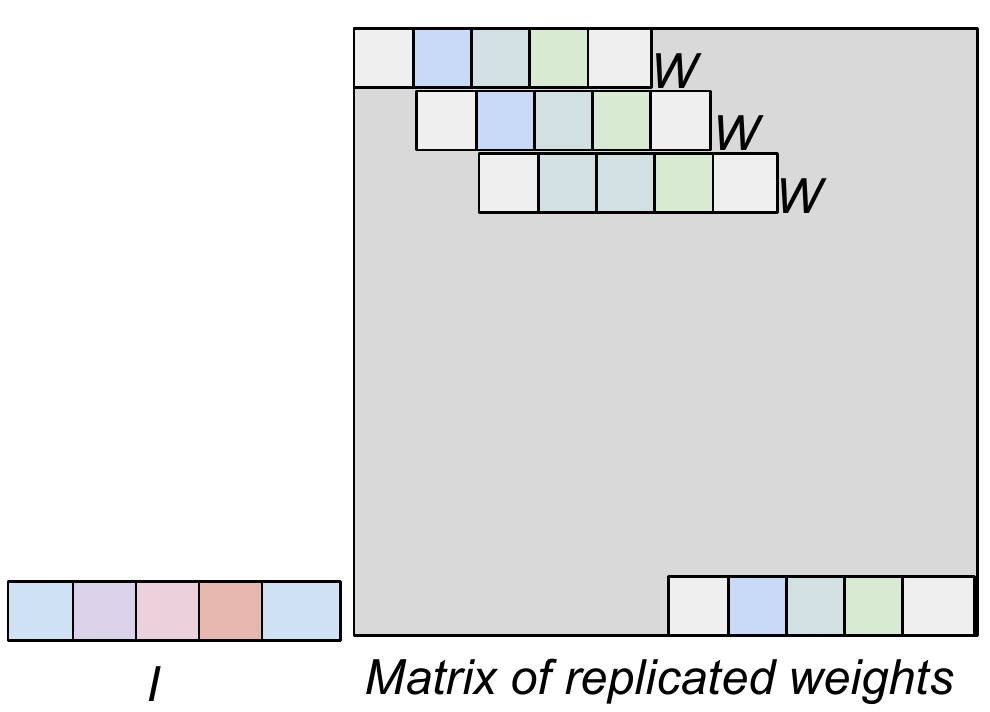}
        \caption{Convolution operation as a matrix operations.}
        \label{fig:gconvOp2}
    \end{minipage}
\end{figure}

The gradient of Equation~\ref{eqn:convOp} wrt. the input takes the form 
\begin{align}
    \frac{\partial o}{\partial i} &=  \frac{\partial i * W}{\partial i}.
\end{align}
Note that gradient of the convolution is a convolutional itself. 
This can be seen in Figure~\ref{fig:gconvOp1}.
We can re-express the {\em convolutional transpose} 
\begin{align}
    \tilde{o} &= \tilde{i} \star W
\end{align}
where $\star$ is a convolutional transpose operator, and $\tilde{o}$ and 
$\tilde{i}$ are just input and output of the function. 

From Figure~\ref{fig:gconvOp2}, we can observe that the gradient of convolutional
formula in Equation~\ref{eqn:convOp} is just a transpose of the replicated input 
matrix. 

Since the convolutional gradient uses the convolutional transpose operator,  
the convolutional transpose can be implemented by using the gradient. 

%Here, we described convolutional tranpose via explaining the gradient of convolutional operator, since the convolutional gradient uses convolutional transpose operator. However, convolutional transpose is its own independent operator.

\begin{figure}[htp]
    \begin{minipage}{0.23\textwidth}
        \includegraphics[width=\columnwidth]{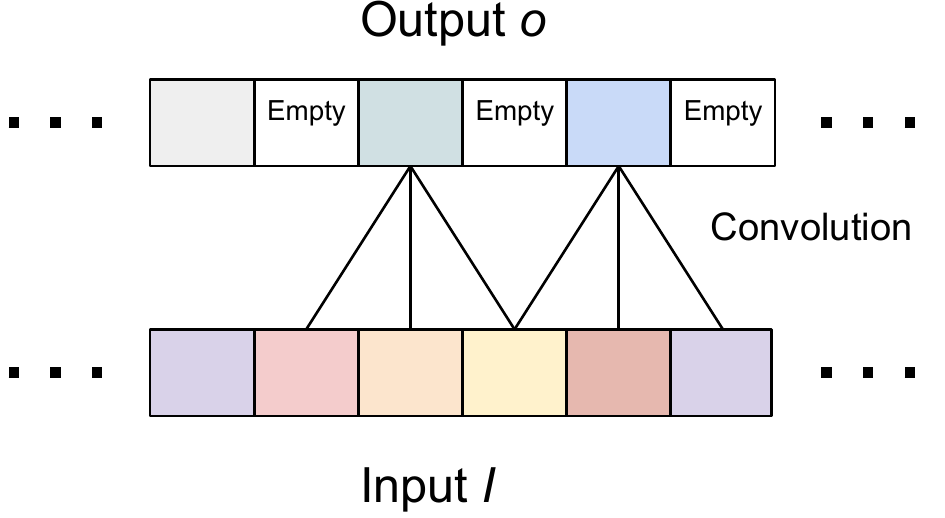}
        \caption{Applying convolution at index $j$.}
        \label{figs:sconvOp1}
    \end{minipage}
    \begin{minipage}{0.24\textwidth}
        \includegraphics[width=\columnwidth]{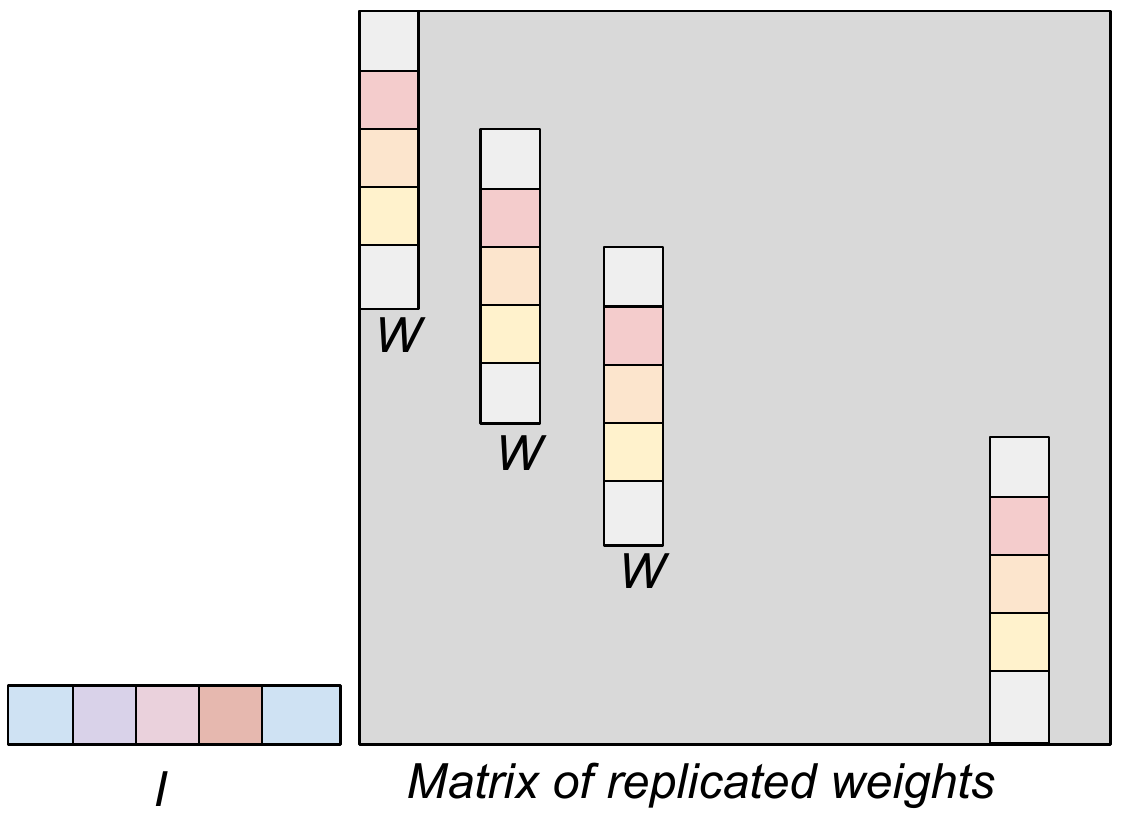}
        \caption{Stride Convolution operation as a matrix operations.}
        \label{figs:sconvOp2}
    \end{minipage}
\end{figure}

Now we consider the case of strided convolution.
For simplicity, we assume that the stride is 2.
Similarly to before, we can write 
\begin{align}
    o = i * W,
    \label{eqn:convOp}
\end{align}
where $i$ is an input, $o$ is an output, and $*$ is the convolution operator.
Figure~\ref{figs:sconvOp1} and Figure~\ref{figs:sconvOp2} illustrate 
the 1D convolution.

Figure~\ref{figs:sconvOp1} shows the visualization of 
2-stride convolution over the input
centered at index $j$ and Figure~\ref{figs:convOp2} presents stride convolution operation
in terms of a matrix operation. 

\begin{figure}[htp]
    \begin{minipage}{0.23\textwidth}
        \includegraphics[width=\columnwidth]{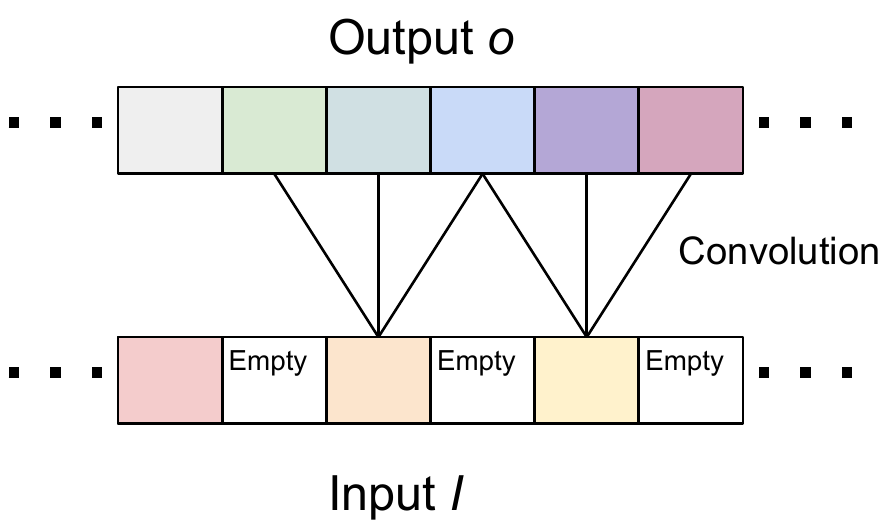}
        \caption{The gradient of convolution at index $k$.}
        \label{fig:sgconvTrOp1}
    \end{minipage}
    \begin{minipage}{0.24\textwidth}
        \includegraphics[width=\columnwidth]{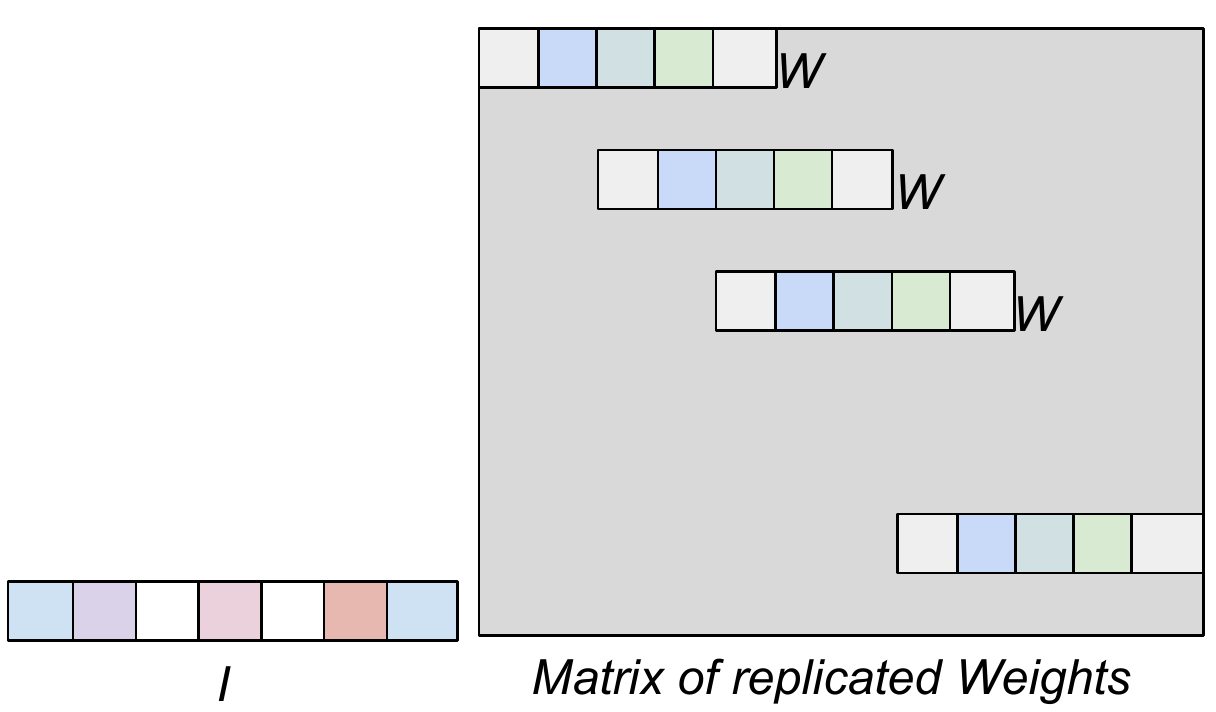}
        \caption{Convolution operation as a matrix operations.}
        \label{fig:sgconvTrOp2}
    \end{minipage}
\vskip -0.2in
\end{figure}

The gradient of Equation~\ref{eqn:convOp} takes the form 
\begin{align}
    \frac{\partial o}{\partial \hat{i}} &= \frac{\partial \hat{i} * W}{\partial i}.
\end{align}
where $\hat{i}$ is the upsampled input $i$ such that $\hat{i}=[i_1, 0, i_2, 0, i_3, \cdots, i_M, 0]$
for stride size equal to 2. Thus, the gradient of the strided convolution is
a convolutional operation on an upsampled version of the input $i$. 
This can be observed from Figure~\ref{fig:sgconvTrOp1}. 

Overall, the convolutional transpose with strides is expressed as 
\begin{align}
    \hat{o} &= \hat{i} \star W
\end{align}
where $\star$ is a convolutional transpose operator, and $\hat{o}$ and 
$\hat{i}$ are just input and output of the function. 
\newpage
\begin{figure}[t]
    \includegraphics[width=\columnwidth]{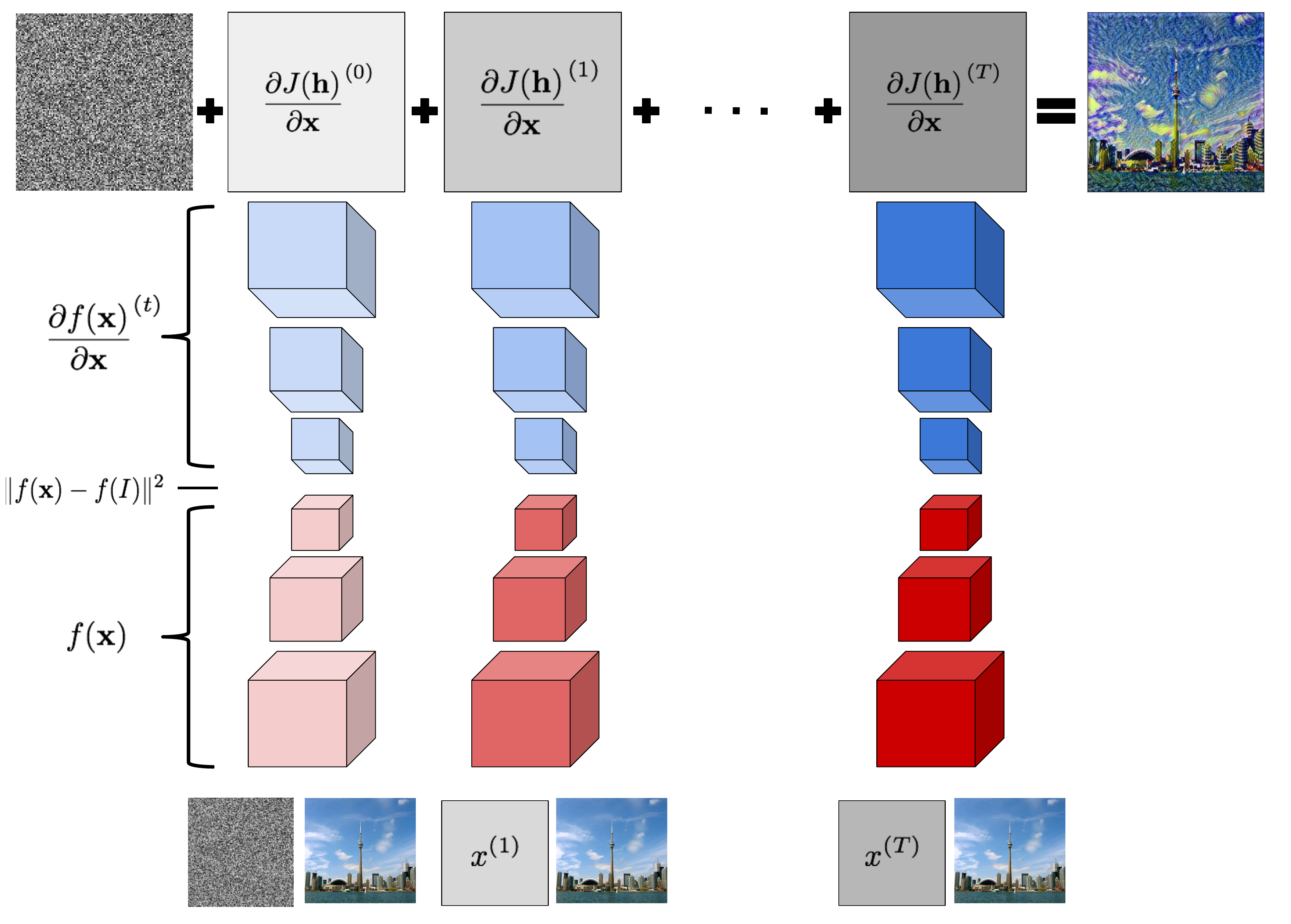}
    \vskip -0.05in
    \caption{The gradient of convolution at index $k$.}
    \label{fig:artify}
\end{figure}
%\begin{figure*}[t]
%    \begin{minipage}{0.49\textwidth}
%        \includegraphics[width=\columnwidth]{Artify2.pdf}
%        \vskip -0.1in
%        \caption{The gradient of convolution at index $k$.}
%        \label{fig:artify}
%    \end{minipage}
%    \begin{minipage}{0.49\textwidth}
%        \includegraphics[width=\columnwidth]{DRAW4.pdf}
%        \vskip -0.05in
%        \caption{The abstract view of DRAW architecture is delineated.}
%        \label{fig:draw}
%    \end{minipage}
%\end{figure*}
%\pagebreak
\subsection*{Relation between sequential modeling and backpropagation with respect to the input methods}

We describe in more detail the relation between sequential modeling algorithms and 
backpropagation with respect to the input (BI). We will 
investigate their relation by examining a specific example.

The goal of BI is to find an optimal input by backpropagating an objective function 
with respect to the input.  
Let $J(\mathbf{x})$ be a differentiable objective function that takes
an input $\mathbf{x}$. 
Then, depending on whether the objective function is non-linear function or not,
we iterate %will obtain an optimal or sub-optimal input by Euler's 1st-order method:
\begin{align}
    x_{opt} = x^{(0)} - \sum_{t=1}^{T} \eta_t \frac{\partial J^{(t)}}{\partial \mathbf{x}^{(t-1)}}
    \label{eqn:BTTI_rule}
\end{align}
where $t$ denotes time, and the chain rule yields %backpropagation with respect to the input is expressed as 
\begin{align}
    \frac{\partial J^{(t)}}{\partial \mathbf{x}^{(t-1)}} 
    = \frac{\partial J^{(t)}}{\partial f^{(t)}} \frac{\partial f^{(t)}}{\partial \mathbf{x}^{(t-1)}},
\end{align}
where $f(\cdot)$ is an intermediate function, which can be composed of many 
non-linear functions like neural networks,
in the objective function $J(\cdot)$.

An example method that uses backpropagation with respect to the input is
%{\em a neural algorithm of artistic style} (Artify) 
\cite{gatys2015neural}. 
We will consider with only the content based objective of this method, 
and compare it to one of the well-known sequential generators, DRAW \cite{Gregor2015}. 
Figure~\ref{fig:artify} presents unrolled version. %of artify\footnote{Note that Figure~\ref{fig:artify} is solely for illustration of our main point.  {\em The style reference image} is not shown in the Figure due to irrelavance of our main message.}. 
The objective function of BI is defined as $\|f_{\mathbf{x}} - f_{I}\|^2$ 
where $f_{\mathbf{x}}$ is the hidden representation of the 
input $\mathbf{x}^{(t)}$, and $h_{I}$ is the 
hidden representation of the {\em reference content image}
of the convolutional network $f(\cdot)$. The network layers are shown 
as red blocks. 
Furthermore, the blue blocks (or the upper half) the diagram in 
Figure~\ref{fig:artify} is the unrolled part of
backpropagation gradient with respect to 
the input $\frac{\partial f}{\partial \mathbf{x}}$.

\begin{figure}[t]
    \includegraphics[width=\columnwidth]{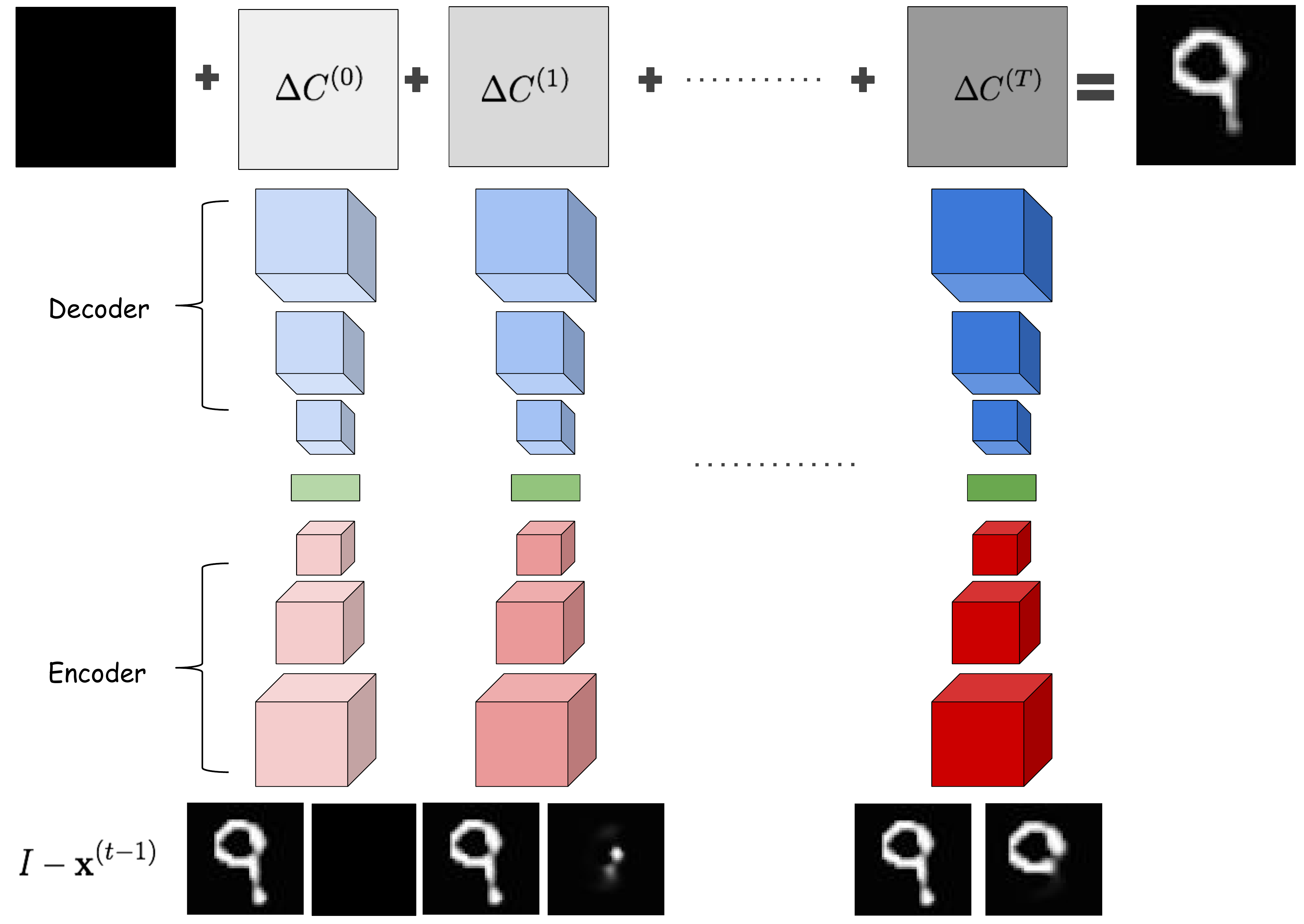}
    \vskip -0.05in
    \caption{The abstract view of DRAW architecture is delineated.}
    \label{fig:draw}
    \vskip -0.05in
\end{figure}

The architecture of DRAW is shown in Figure~\ref{fig:draw}\footnote{The attention
mechanism is omitted for clarity.}. 
DRAW takes the input and the difference between the input and canvas at time $t$, 
and it propagates through the encoder and decoder. 
At the end of each time step, DRAW outputs the updated canvas $C^{(t)}$ by
updating the previous canvas $C^{(t-1)}$ with what will be drawn at time $t$, which is 
equivalent to change in the canvas $\Delta C^{(t)}$. 

We can immediately notice the similarity between two architectures. 
The update procedure of draw, which is expressed as 
\begin{align}
    C^{(t)} = C^{(t-1)} + \Delta C^{(t)},
\end{align}    
resembles the update rule of BI in Equation~\ref{eqn:BTTI_rule}. Moreover,
the encoder of DRAW, $enc(\cdot)$, can be seen as some function $f(\cdot)$, which will be the
convolutional neural network in BI. 
Similarly, the decoder of DRAW, $dec(\cdot)$, can be seen as 
the unrolled version of BI, which corresponds 
to $\frac{\partial f^{(t)}}{\partial \mathbf{x}^{(t-1)}}$.
The main difference is that BI takes the difference
in the hidden representation space of $f(\cdot)$ and DRAW takes the difference
from the original input space.

Overall, we linked each components of two models by examining the abstraction 
as shown:
\begin{list}{\labelitemi}{\leftmargin=1em}
    \vspace{-0.25cm}
    \itemsep0.em 
    \item $\Delta C^{(t)}$ reflects $\frac{\partial J^{(t)}}{\partial \mathbf{x}^{(t-1)}}$.
    \item $enc(\cdot)$ and $dec(\cdot)$ reflect $f(\cdot)$ and $\frac{\partial f^{(t)}}{\partial \mathbf{x}^{(t-1)}}$.
\end{list}
%\begin{wrapfigure}{r}{0.4\textwidth}
\begin{figure}
\begin{minipage}{0.4\textwidth}
\begin{algorithm}[H]
\begin{algorithmic} 
\STATE Initial hidden state: $\mathbf{h}_{c,0}=\bm{0}$.
\WHILE{$t < T$}
\STATE $\bz_t \sim p(Z)$
\STATE $\mathbf{h}_{z} = \tanh(W \bz_t + \mathbf{b})$ 
\STATE $\mathbf{h}_{c,t} = g(\Delta C_{t-1})$
\STATE $\Delta C_t = f([\mathbf{h}_{z}, \mathbf{h}_{c,t}])$
\ENDWHILE
\STATE $\mathcal{C} = \sigma(\sum^{T}_{t=1} \Delta C_t)$.
\end{algorithmic} 
\caption{GRAN's sample generating process.}% $f$ and $g$ are the encoding and decoding network.}
\end{algorithm}
\end{minipage}
%\end{wrapfigure}
\end{figure}

\pagebreak

\begin{figure}[t] 
%\begin{wrapfigure}{r}{0.5\textwidth}                                        
        \vskip -0.25in
        \centering
        \includegraphics[width=\columnwidth]{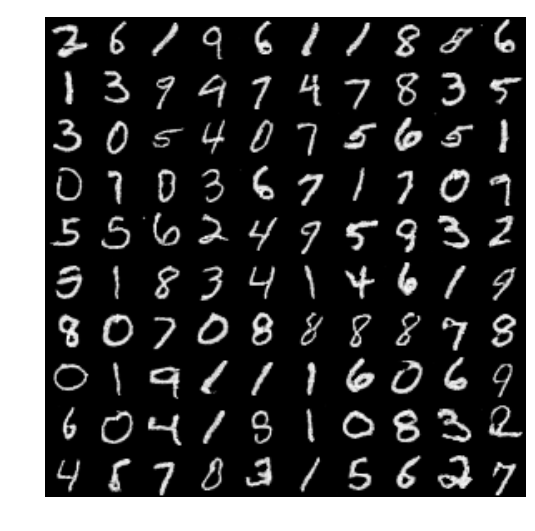}
        \vskip -0.1in
        \caption{MNIST Samples generated by GRAN}
        \label{figs:mnist_samples}                                               
%\end{wrapfigure}                   
\end{figure}

\subsection*{Supplementary Material to the Experiments}
\label{sup:exp}

\textbf{The MNIST dataset} contains 60,000 images for training and 10,000 images for testing and each of the images is
28$\times$28 pixels for handwritten digits from 0 to 9 (LeCun et al., 1998). Out of the 60,000 training
examples, we used 10,000 examples as validation set to tune the hyper-parameters of our model.\\
\textbf{The CIFAR10 dataset} consists of 60000 32$\times$32 colour images in 10 classes,
with 6000 images per class. There are 50000 training images and 10000 test images. \\
\textbf{The LSUN Church dataset} consists of high resolution natural scene images with 10 different
classes \cite{Yu2015}. We considered training on outdoor church images, which contains 126,227 training,
 300 validaiton, and 1000 test images. These images were downsampled to 64$\times$64 pixels.\\
\textbf{The LSUN Living Room + Kitchen dataset} consists of high resolution natural scene images with 10 different
classes \cite{Yu2015}. We considered training on living room and kitchen images,
which contains approx. 120,000 training, 300 validaiton, and 1000 test images. 
These images were downsampled to 64$\times$64 pixels.  

\begin{figure}
%\begin{wrapfigure}{l}{0.5\textwidth}
    \vskip -0.1in
    \begin{minipage}{0.49\textwidth}
        \centering
        \includegraphics[width=\columnwidth]{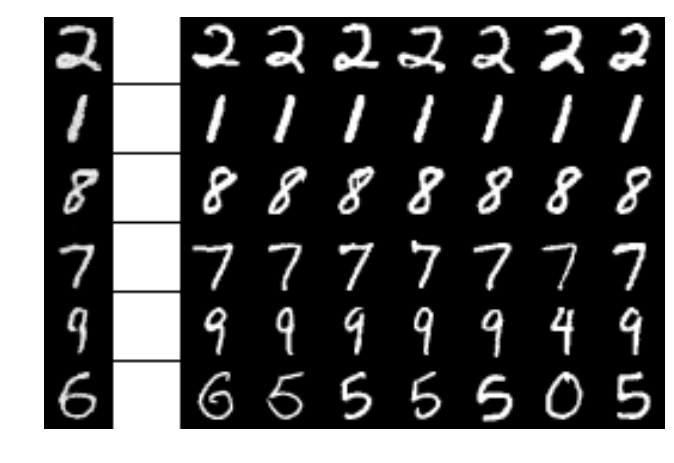}
        \vskip -0.1in
        \caption{Nearest Neighbour training examples for MNIST samples.}
        \label{figs:mnist_NN_samples}
    \end{minipage}
    \begin{minipage}{0.49\textwidth}
        \centering                                                              
        \includegraphics[width=\columnwidth]{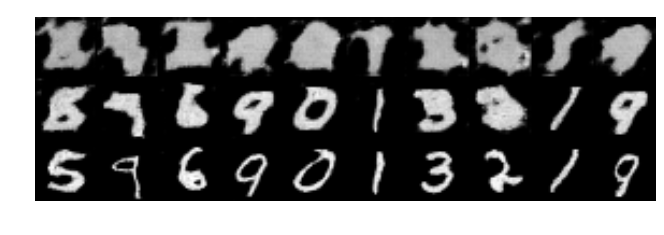} 
        \vskip -0.1in                                                           
        \caption{Drawing at different time steps on mnist samples.}             
        \label{figs:mnist_seq}                                                  
    \end{minipage} 
    \vskip -0.1in                                                           
%\end{wrapfigure}
\end{figure}
All datasets were normalized such that each pixel value ranges in between [0, 1].
For all of our results, we optimized the models with ADAM \cite{Kingma2015}. 
The batch size was set to 100, and the learning rate was selected from a discrete range
chosen based on the validation set. Importantly, we used different learning rates
for the discriminative network and generative network. 
Throughout the experiments, we found that having different learning rates are useful to obtain 
succesfully trained generative adversarial models.
As our proposed model is a sequential generator, we must select the number 
of steps, $T$, to run the model for generation. 
We compared the models in different number of timesteps,
$\lbrace 1,3,5\rbrace$. 
Note that GRAN is equivalent to DCGAN when $T=1$ up to
one extra fully connected layer. We denote this as GRAN1.

%\begin{figure*}[t]
%    \centering
%    \includegraphics[width=0.9\textwidth]{GRAN0.pdf}
%    \vskip -0.1in
%    \caption{Depiction of single time step component of Generative Recurrent Adversarial Networks architecture layed out. 
%    (The numbers of the figures are used for modelling CIFAR10 dataset)}
%    \label{figs:GRAN}
%\vskip -0.2in
%\end{figure*} 
Throughout the experiments, we used a similar architecture for the generative and 
discriminative network as shown in Figure~\ref{figs:GRAN} and 
Figure~\ref{figs:GRAN_skeleton}. 

%\begin{wrapfigure}{l}{0.66\textwidth}
\begin{table}[htp]
    \vskip -0.05in
    \caption{The experimental hyper-parameters on different data sets.}
    \label{tab:exp_hyper}
    \begin{small}
    \begin{sc}
    \begin{tabular}{lcccc}
    \hline
    %\abovespace\belowspace
    Dataset & \# Kernels & Filter Sz. & \# $\bz$ \\
    \hline
    %\abovespace
    MNIST     & [80, 40, 1]         & [5,5,5]   & 60\\%& $\surd$  & 100 \\
    CIFAR10   & [1024, 512, 216, 3] & [5,5,5,5] & 100\\%& $\times$ & 100 \\
    %\belowspace
    LSUN   & [1024, 512, 216, 128, 3] & [5,5,5,5,5] & 100\\%& $\times$ & 100 \\
    \hline
    \end{tabular}
    \end{sc}
    \end{small}
    \vskip -0.05in
\end{table}
%\end{wrapfigure}

The convolutional layers of the networks for the discriminator and generator 
are shared. % since they both take the same dimension as the input.
The number of convolutional layers and the number of hidden units 
at each layer were varied based on the dataset. Table~\ref{tab:exp_hyper} shows
the number of convolution kernels and the size of the filters at each convolutional layer. 
The numbers in the array from left to right corresponds to each bottom to top 
layer of the convolutional neural network.
%In order to design experiments such that the same nubmer of stochasity is injected for GRAN and other
%GAN types of models, we secregated the latent variable $\bz$ to $T$ many $\bz_i$ variables. 
%For example, if $|\bz|=100$ for DCGAN, then we used latent variables 
%$\forall \ i=1,\cdots,T, \ |\bz_i|=\frac{|\bz|}{T} = 20$ for GRAN with the time length $T=5$.
One can tie the weights of convolution
and convolutional transpose in the encoder and decoder of GRAN 
to have the same number of parameters for both DCGAN and GRAN.

The quality of samples can depend on several tricks \cite{Radford2015}, including:
\begin{enumerate}[leftmargin=0.5cm]
    \item Removing fully connected hidden layers and 
        replacing pooling layers with strided convolutions on the discriminator \cite{Springenberg2014}
        and fractional-strided convolutions (upsampling) on the generator.
    \item Using batch-normalization \cite{Ioffe2015} on both generative and discriminative models.
    \item Using ReLU activations in every layer of the generative model except the last layer, and using
        LeakyReLU activations \cite{Maas2013} in all layers of the discriminative model.
\end{enumerate}
Overall, these architectural tricks make it easier to generate 
smooth and realistic samples. We rely on these tricks 
and incorporate them into GRAN.

%\begin{wrapfigure}{r}{0.5\textwidth}
\begin{figure}[h]
    \begin{minipage}{0.49\textwidth}
        \centering
        \includegraphics[width=0.5\textwidth]{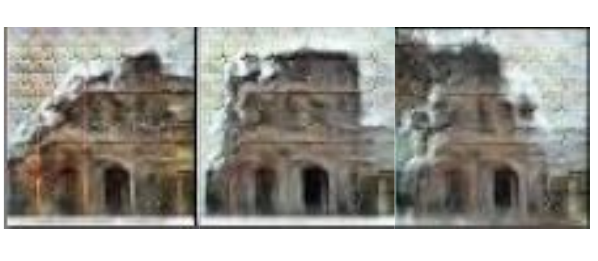}
        \vskip -0.05in
        \caption{Example of three different churches (samples) with some similarities.}
        \label{figs:cifar10_seq}
    \end{minipage}
\end{figure} 
%\end{wrapfigure} 

\textbf{Analysis on GRAN samples} 
The following figure is to support the studies in the experiments,
particularly it supports the description under
\textit{Q: Does GRAN overfit the training
data?}

\newpage 
\textbf{Nearest Neighbours of samples from the training dataset}
%\begin{wrapfigure}{r}{0.5\textwidth}
\begin{figure}[htp]
    \vskip -0.1in
    \begin{minipage}{0.49\textwidth}
        \centerline{\includegraphics[width=\columnwidth]{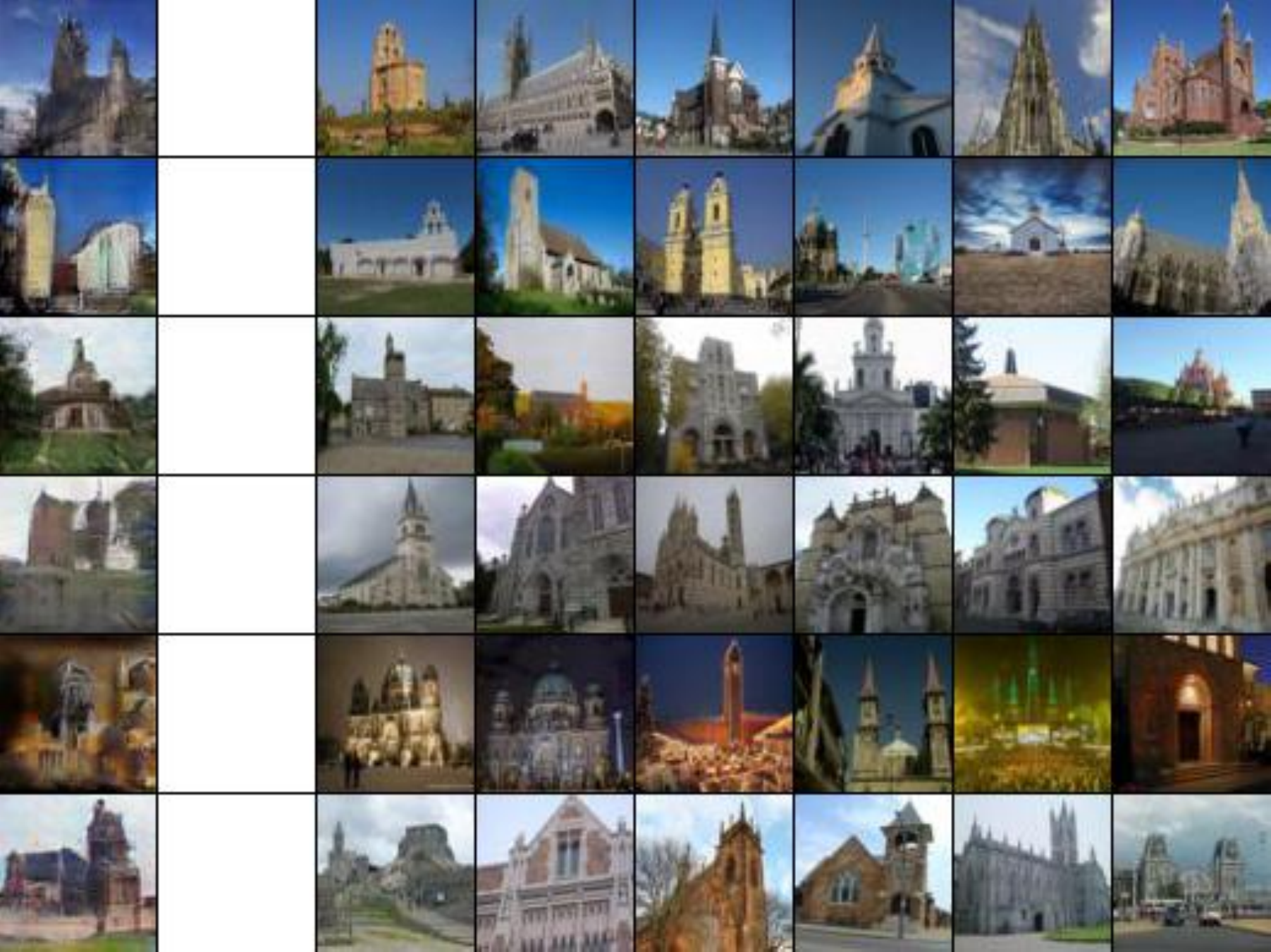}}
        \vskip -0.05in
        \caption{Nearest Neighbour training examples for lsun churchsamples using GRAN5.}
        \label{figs:cifar10_seq3}
    \end{minipage}
    \begin{minipage}{0.49\textwidth}
        \centerline{\includegraphics[width=\columnwidth]{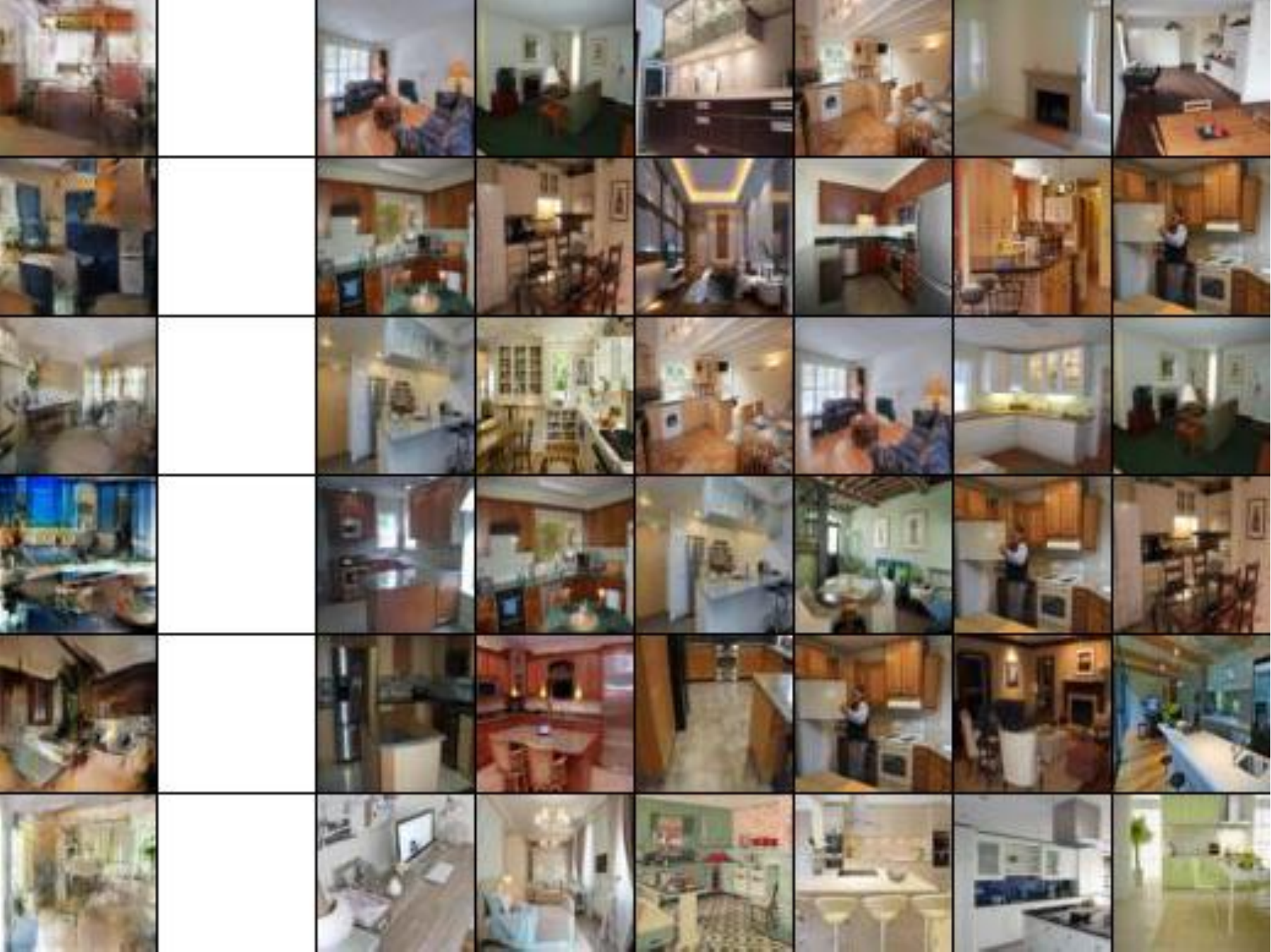}}
        \vskip -0.05in
        \caption{Nearest Neighbour training examples for lsun (living room + kitchen) samples.}
        \label{figs:cifar10_NN_samples} 
    \end{minipage}
    \vskip -0.1in
\end{figure}
%\end{wrapfigure} 

%\textbf{Samples when injecting different noises at each time step}
%\begin{figure}[htp]
%    \begin{minipage}{0.49\textwidth}
%        %\includegraphics[width=\columnwidth]{cifar10_samples1.pdf}
%        \includegraphics[scale=0.9,width=\columnwidth]{cifar10_gran_samples1.pdf}
%        \vskip -0.1in
%        \caption{Cifar10 samples generated by GRAN with injecting different noises at every time step}
%        \label{figs:cifar10_samples_diff_Zs}
%    \end{minipage}
%    \begin{minipage}{0.49\textwidth}
%        %\includegraphics[width=\columnwidth]{lsun_samples.pdf}
%        \includegraphics[width=\columnwidth]{lsun_gran_samples.pdf}
%        \vskip -0.1in
%        \caption{LSUN samples generated by GRAN with injecting different noises at every time step}
%        \label{figs:lsun_samples_diff_Zs}
%    \end{minipage}
%    \vskip -0.15in
%\end{figure}

{\bf Intermediate samples at time step for GRAN5}
\begin{figure}[htp]
    \vskip -0.1in
    \begin{minipage}{0.49\textwidth}
        \begin{center}
        \includegraphics[width=\columnwidth]{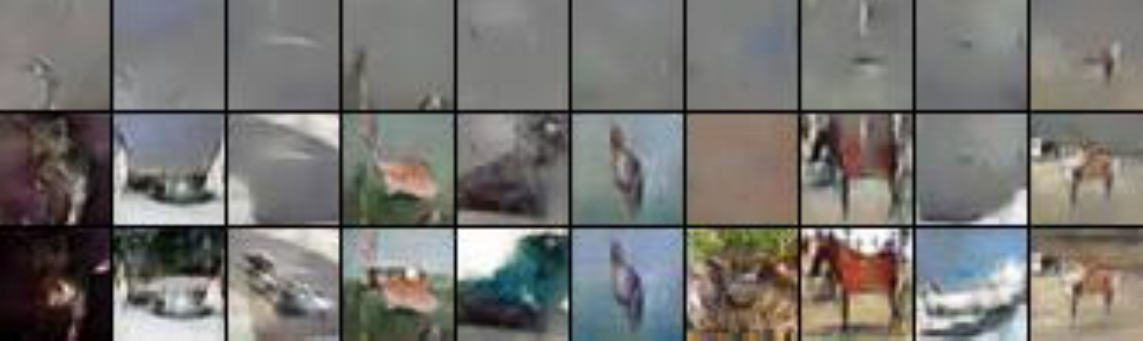}
        \vskip -0.05in
        \caption{Drawing at different time steps on cifar10 samples with injecting different noises at every time step.}
        \label{figs:cifar10_seq_diff_Zs}
        \end{center}
    \end{minipage}
    \begin{minipage}{0.49\textwidth}
        \begin{center}
        \includegraphics[width=\columnwidth]{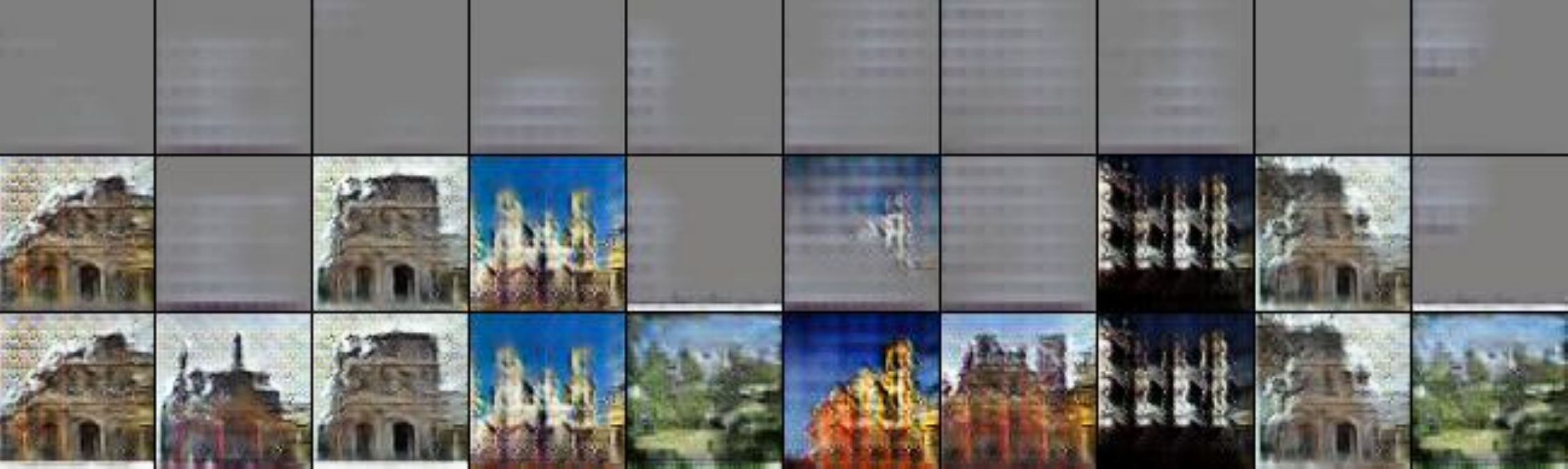}
        \vskip -0.05in
        \caption{Drawing at different time steps on lsun samples with injecting different noises at every time step.}
        \label{figs:lsun_seq_diff_Zs}
        \end{center}
    \end{minipage}\\
    \begin{minipage}{0.49\textwidth}
        \begin{center}
        \includegraphics[width=\columnwidth]{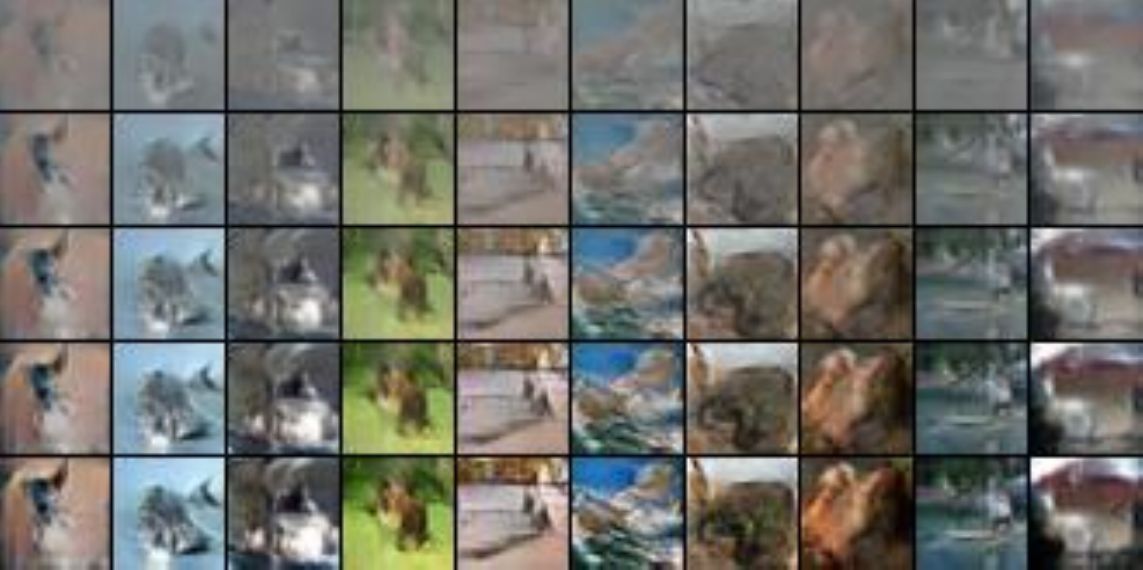}
        \vskip -0.05in
        \caption{Drawing at different time steps on cifar10 samples.}
        \label{figs:cifar10_seq5}
        \end{center}
    \end{minipage}
    \begin{minipage}{0.49\textwidth}
        \begin{center}
        \includegraphics[width=\columnwidth]{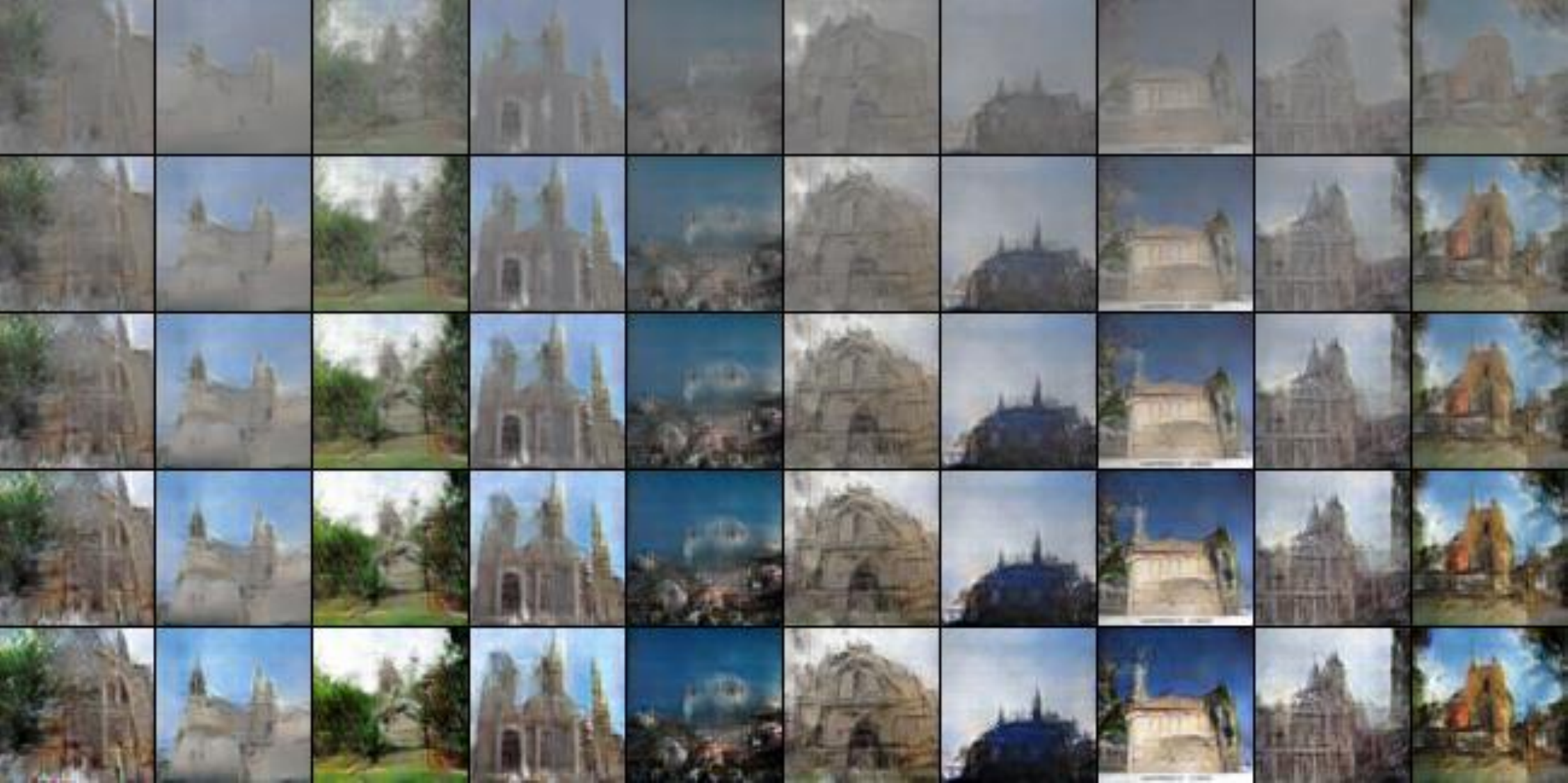}
        \vskip -0.05in
        \caption{Drawing at different time steps on lsun church samples.}
        \label{figs:lsun_seq}
        \end{center}
    \end{minipage}\\
    \begin{minipage}{0.49\textwidth}
        \begin{center}
        \includegraphics[width=\columnwidth]{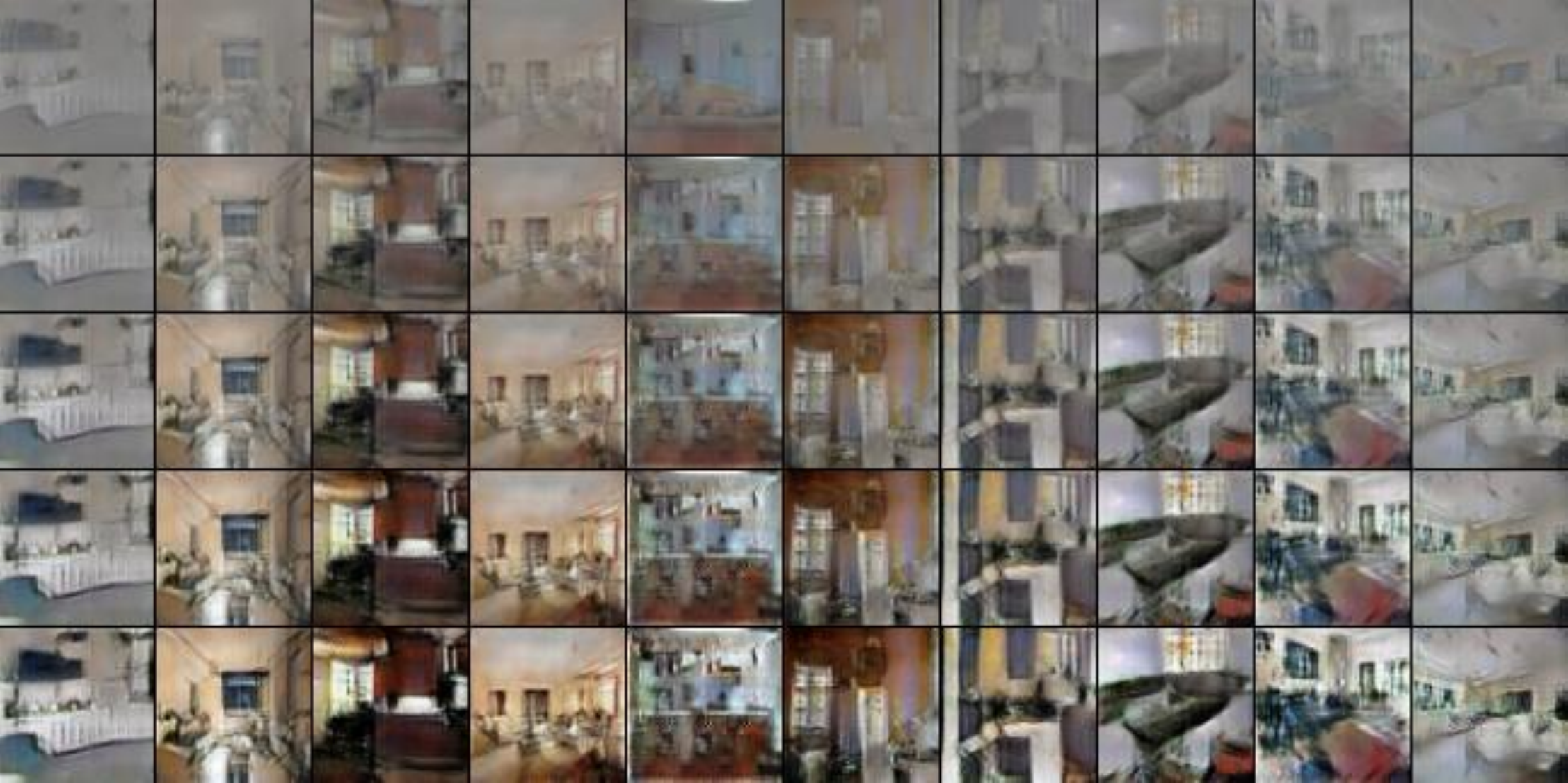}
        \vskip -0.05in
        \caption{Drawing at different time steps on lsun (living room + kitchen) samples.}
        \label{figs:lsun_seq}
        \end{center}
    \end{minipage}\\
\end{figure} 

\begin{figure}[htp]
    \begin{center}
        \centerline{\includegraphics[width=\columnwidth]{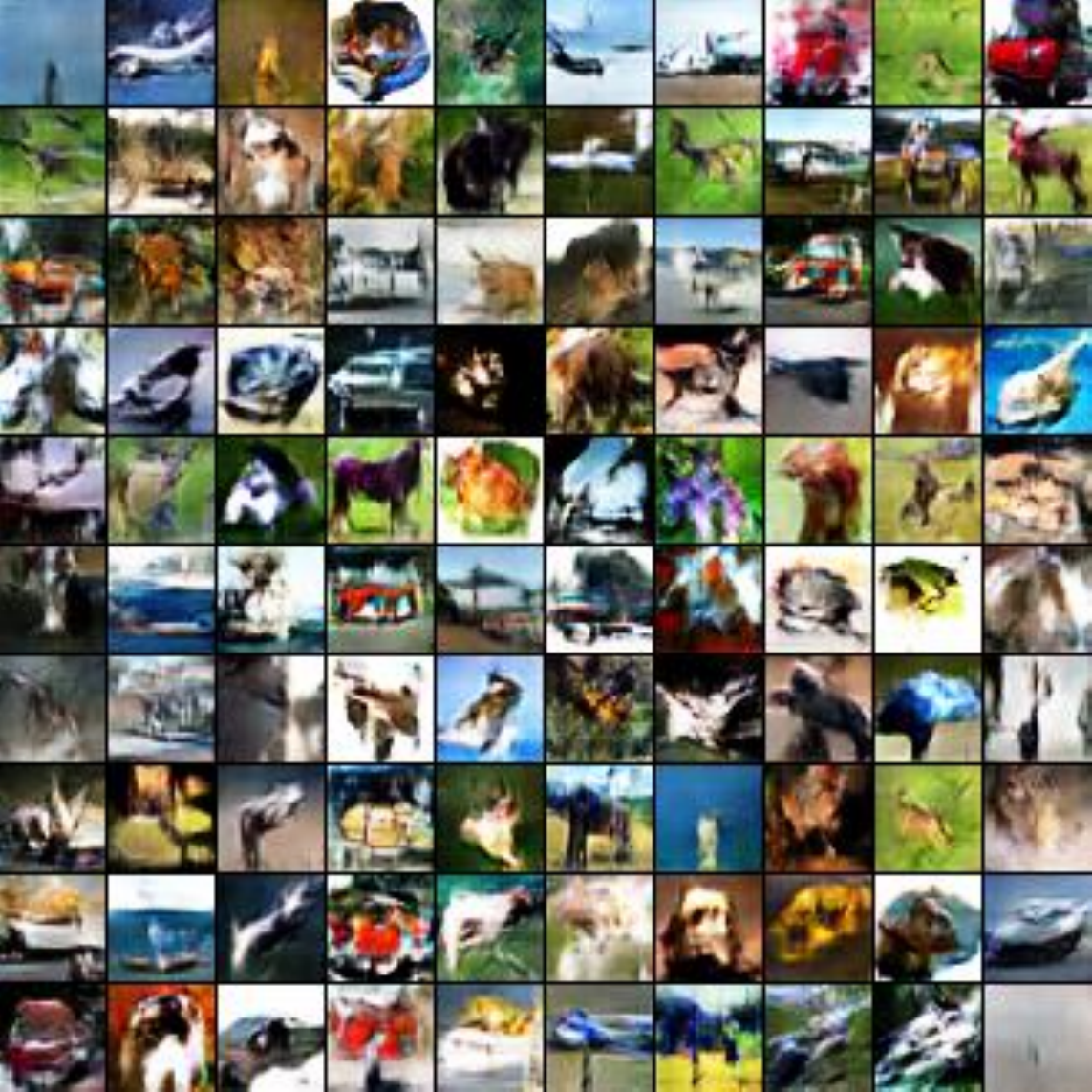}}
        \caption{CIFAR10 Samples generated by GRAN3 with $7$-steps.}
        \centerline{\includegraphics[width=\columnwidth]{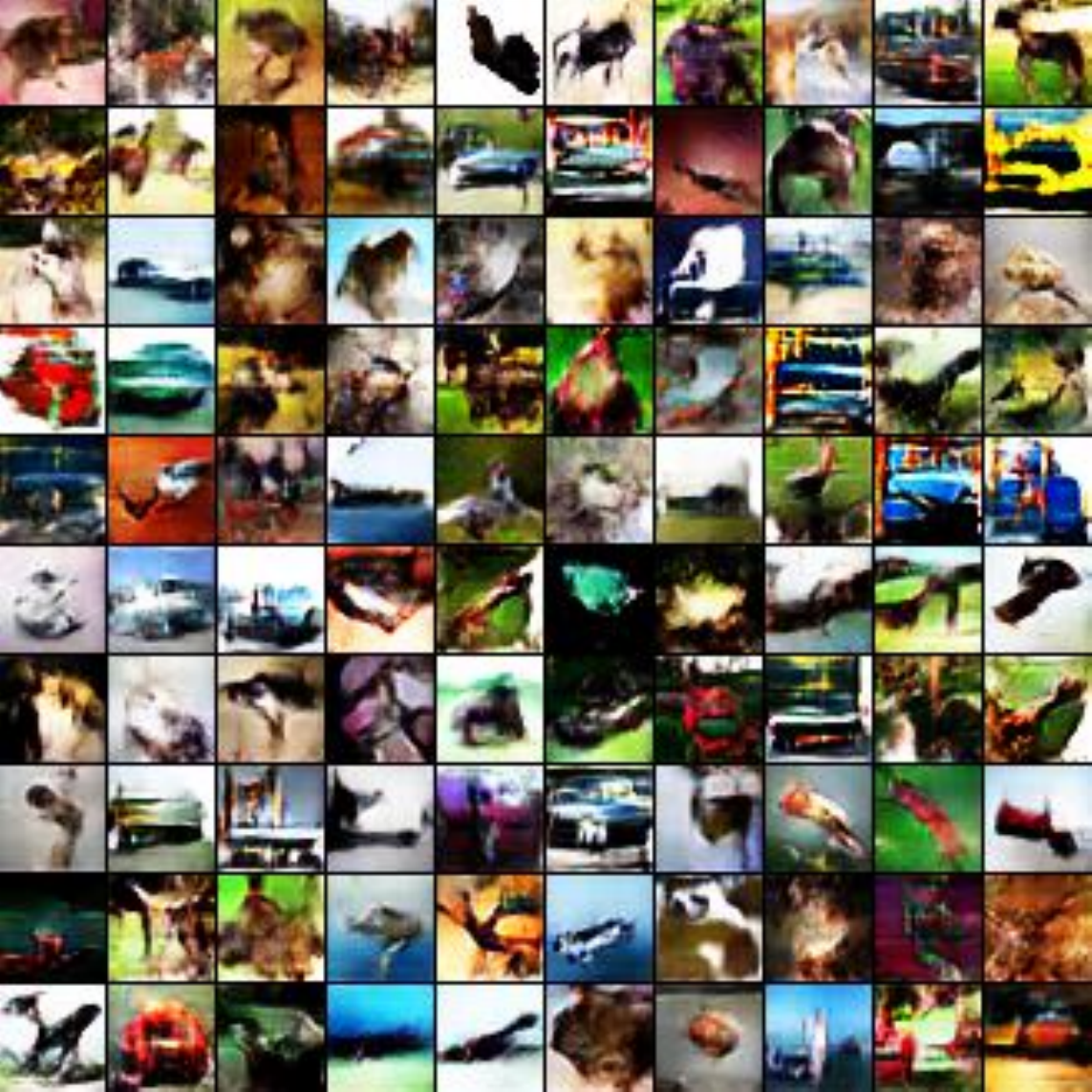}}
        \caption{CIFAR10 Samples generated by GRAN3 with $9$-steps.}
        \label{figs:mult_steps}
    \end{center}
\end{figure}

\newpage
\begin{figure*}[t]
\begin{center}
    \centering
    \includegraphics[width=2\columnwidth]{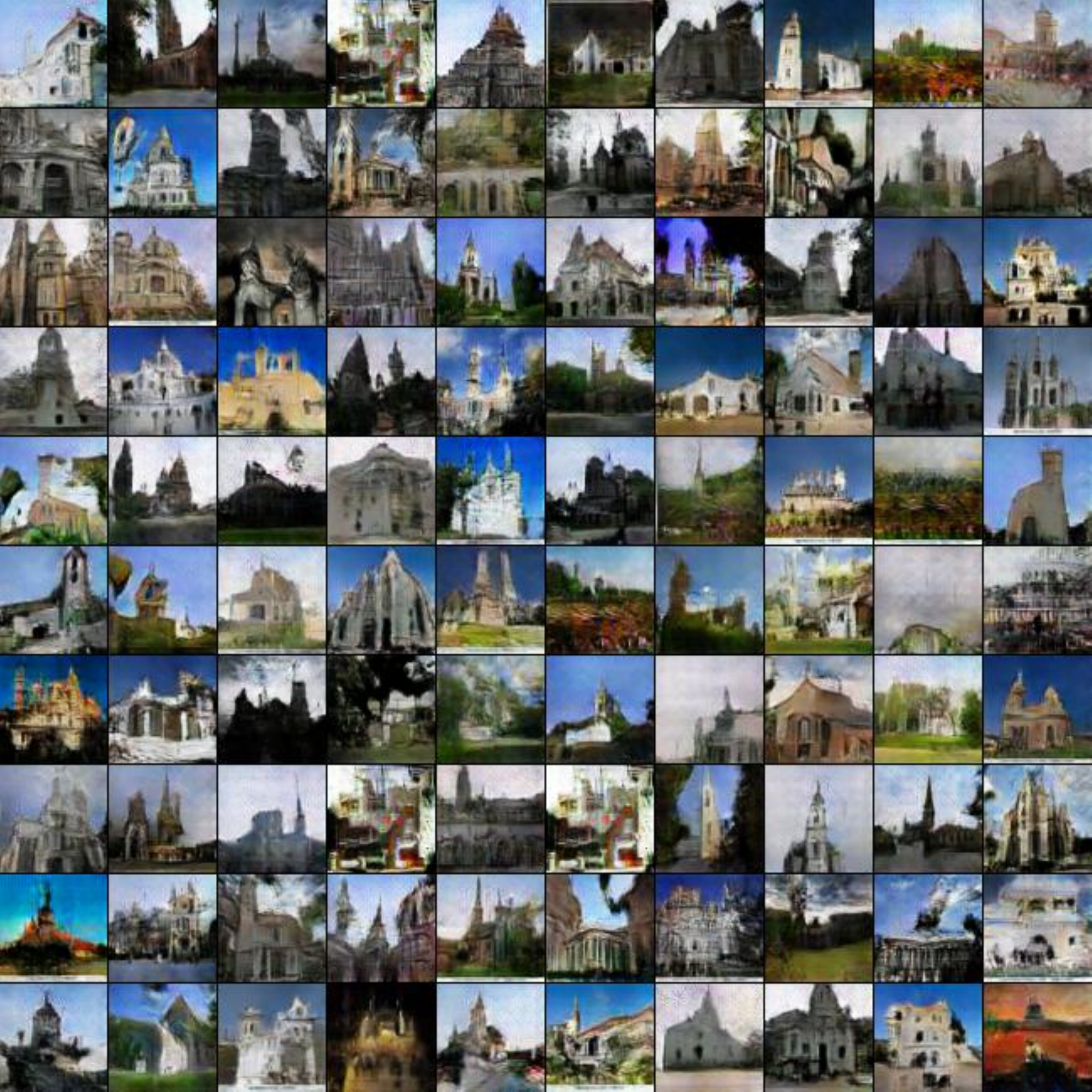}
    \caption{LSUN (church) samples generated by GRAN5.}
\label{figs:lsun_samples_gran5}
\end{center}
\end{figure*} 

\begin{figure*}[t]
\begin{center}
    \centering
    \includegraphics[width=2\columnwidth]{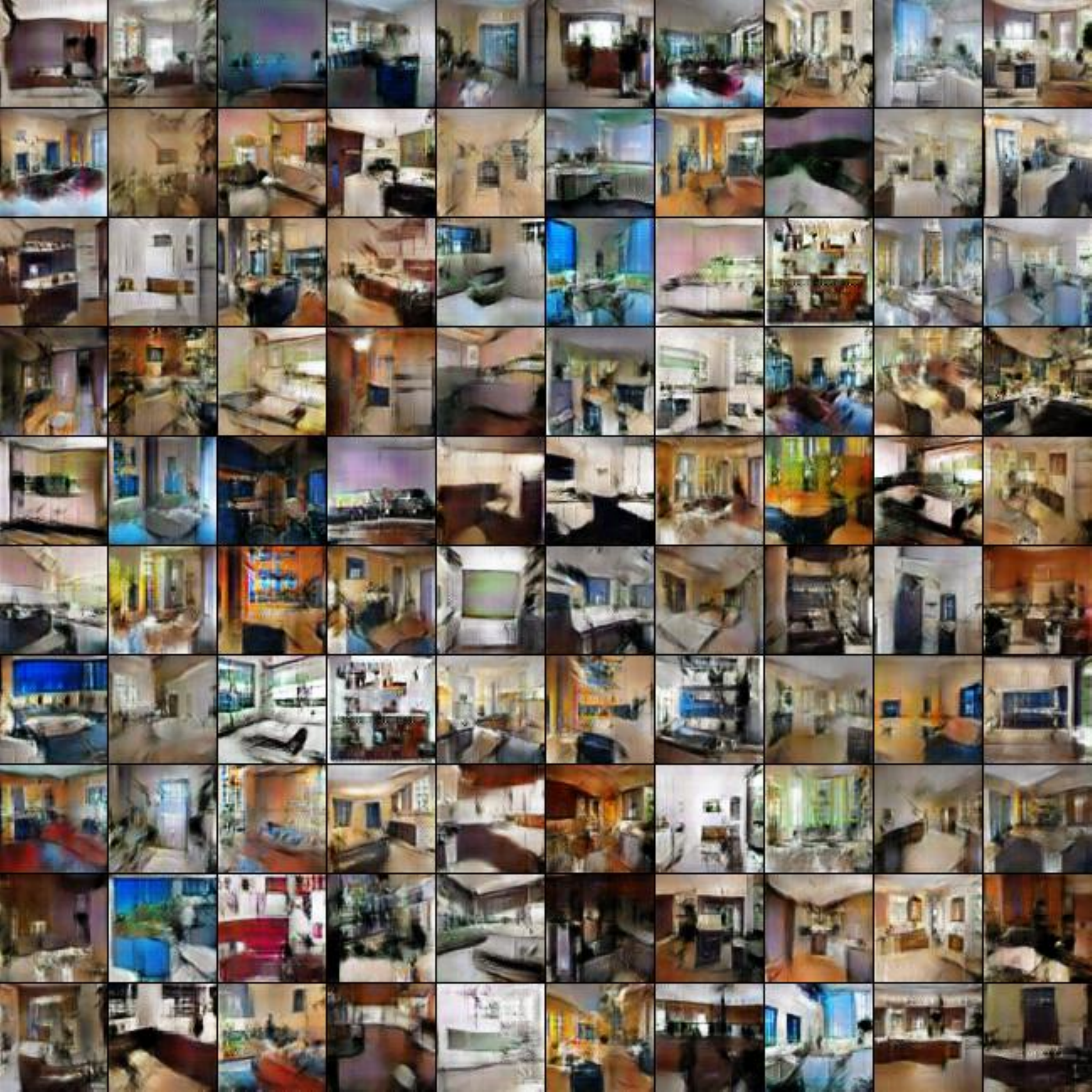}
\caption{LSUN (living room +kitchen) samples generated by GRAN5.}
\label{figs:lsun_samples_lk}
\end{center}
\end{figure*}

\begin{figure*}[t]
\begin{center}
    \centering
    \includegraphics[width=2\columnwidth]{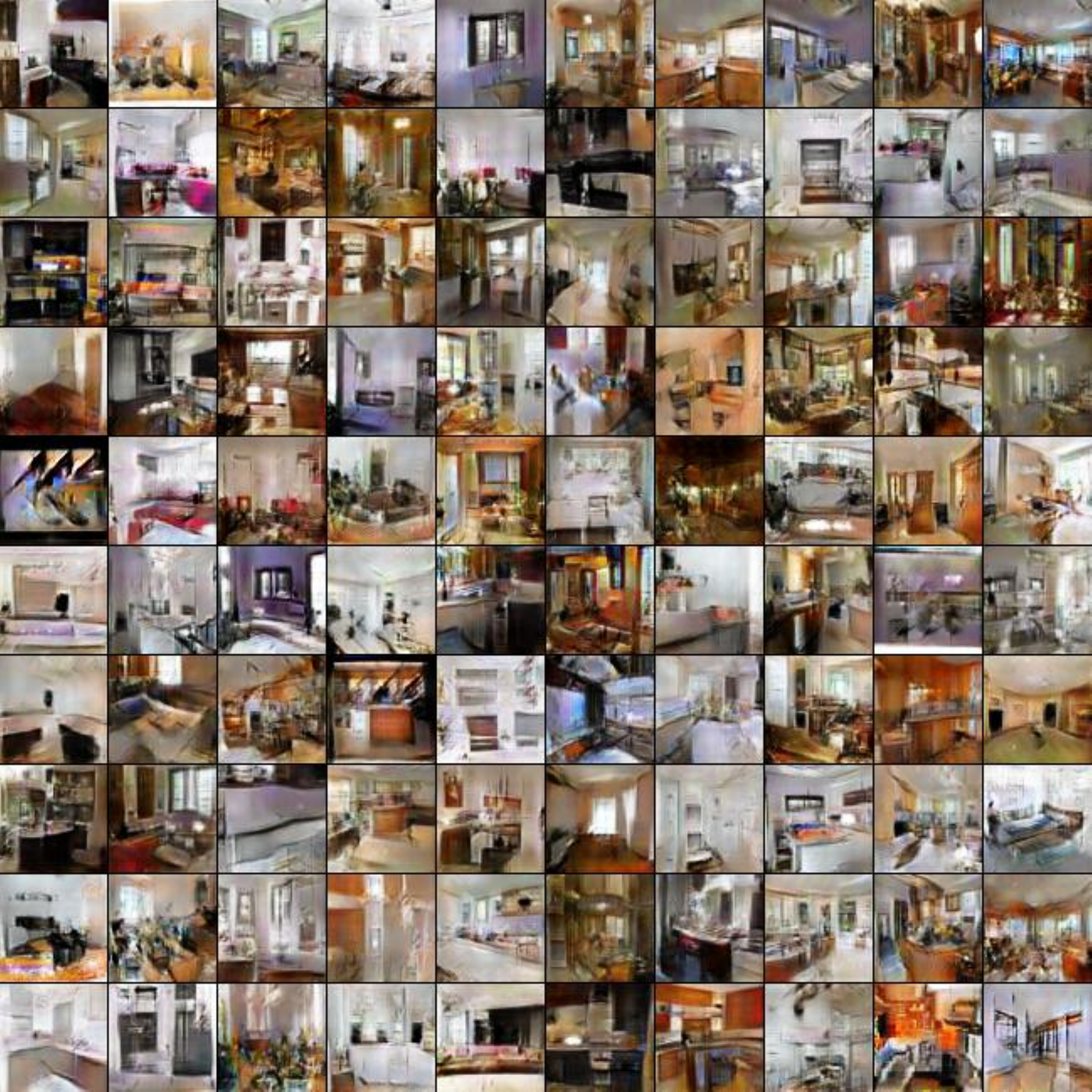}
\caption{LSUN (living room +kitchen) samples generated by GRAN3.}
\label{figs:lsun_samples_gran5}
\vskip -0.1in
\end{center}
\vskip -0.1in
\end{figure*} 

\clearpage
\textbf{ImageNet samples}

We also trained GRAN on ImageNet dataset.
ImageNet dataset (Deng et al., 2009) is a high resolution natural images.
We rescaled the images to 64 $\times$ 64 pixels. The architecture that was used
for ImangeNet is same as the architecture that was used for LSUN datset 
except that there are three times more kerns on both generator and discriminator.

\begin{figure*}[htp]
    \begin{center}
        \centerline{\includegraphics[width=2\columnwidth]{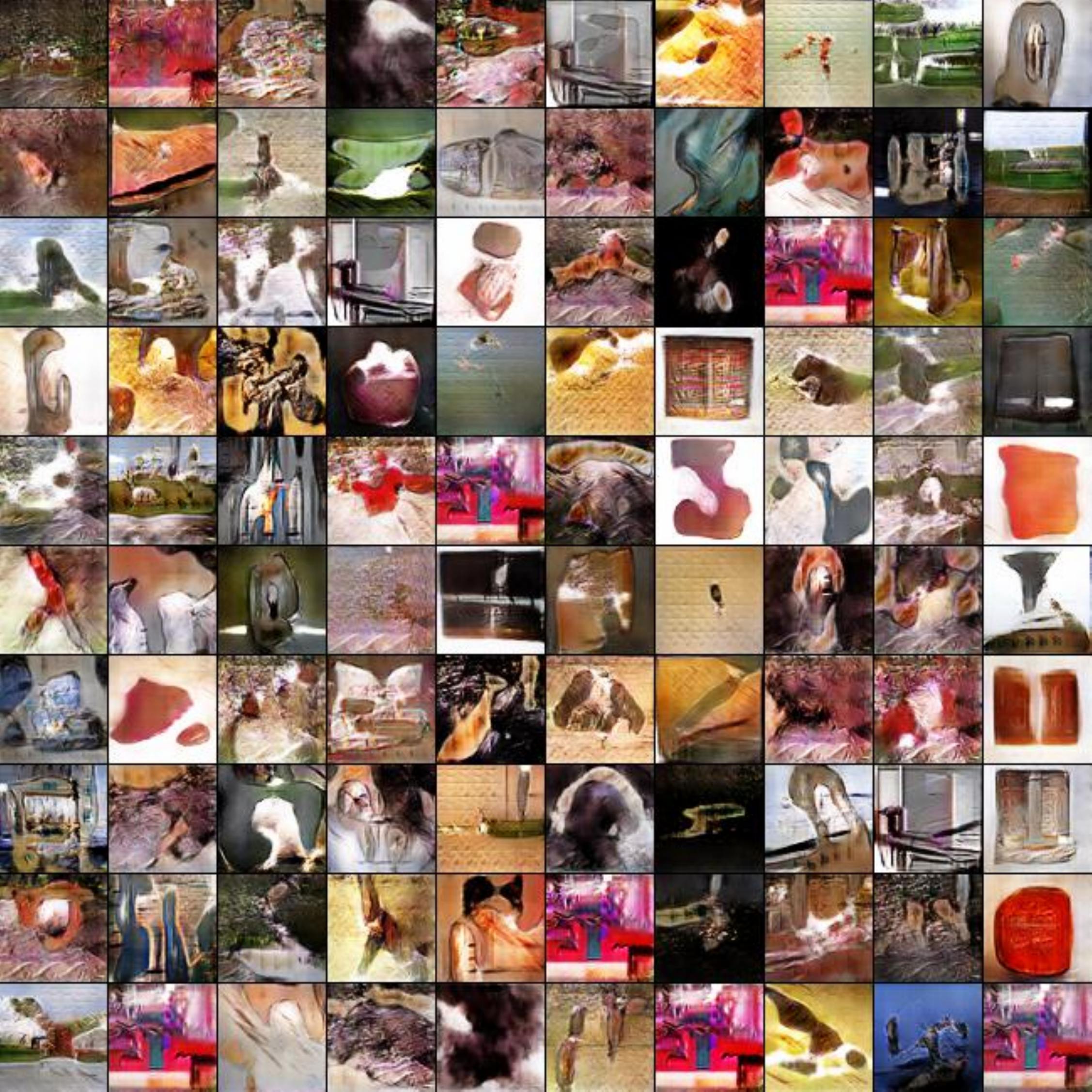}}
        \vskip -0.1in
        \caption{Samples of ImageNet images generated by GRAN3.}
        \label{figs:IN_samples}
        \end{center}
\end{figure*}

The samples are shown in Figure~\ref{figs:IN_samples}. Unfortunately, the samples does not generate objects 
from ImageNet dataset. However, it also shows that they are not overfitting,
because they do not show actual objects but they are quite artistic.
We hypothesize that this is becasue the model does not have the capacity to model
1000 object classes. Hence, they stay as abstract objects.

\end{document}